\documentclass[preprint]{elsarticle}

\usepackage{lineno,hyperref}

\usepackage{amsmath, graphicx, makecell, amsfonts, amssymb, algorithm, algpseudocode}
\usepackage{stmaryrd}
\usepackage{scalerel}
\usepackage{bbm}

\modulolinenumbers[5]

\journal{}
\date{}

\def\x{{\mathbf x}}
\def\L{{\cal L}}
\def\D{{\cal D}}
\def\H{{\cal H}}
\def\R{{\cal R}}
\def\E{{\mathbb E}}
\def\MC{{\scaleto{MC}{3pt}}}
\def\yhat{{\widehat{y}}}

\def\ind{{\mathbbm{1}}}

\begin{document}

\begin{frontmatter}

\title{Unsupervised Domain Adaptation \\Based on the Predictive Uncertainty of Models}

\author[firstaffiliation,secondaffiliation]{JoonHo~Lee}

\author[firstaffiliation]{Gyemin~Lee}

\address[firstaffiliation]{Department of Electronic and IT Media Engineering, Seoul National University of Science and Technology, Republic of Korea}
\address[secondaffiliation]{Machine Learning Research Center, Samsung SDS Technology Research, Republic of Korea}

\begin{abstract}
Unsupervised domain adaptation (UDA) aims to improve the prediction performance in the target domain under distribution shifts from the source domain. 
The key principle of UDA is to minimize the divergence between the source and the target domains. 
To follow this principle, many methods employ a domain discriminator to match the feature distributions.
Some recent methods evaluate the discrepancy between two predictions on target samples to detect those that deviate from the source distribution. 
However, their performance is limited because they either match the marginal distributions or measure the divergence conservatively.
In this paper, we present a novel UDA method that learns domain-invariant features that minimize the domain divergence. 
We propose model uncertainty as a measure of the domain divergence. 
Our UDA method based on model uncertainty (MUDA) adopts a Bayesian framework and provides an efficient way to evaluate model uncertainty by means of Monte Carlo dropout sampling. 
Empirical results on image recognition tasks show that our method is superior to existing state-of-the-art methods. 
We also extend MUDA to multi-source domain adaptation problems.
\end{abstract}


\end{frontmatter}


\section{Introduction}
\label{sec:intro}

Deep neural networks (DNNs) have shown great success in numerous image processing and computer vision tasks thanks to large amounts of well-annotated training data.
However, DNN models often fail to work well in real world applications due to the distribution shift of the target data from the data used to train the models.
Collecting new large-scale labeled data for the target task is prohibitively expensive and time-consuming in many cases.

Domain adaptation (DA) addresses this \textit{domain shift} problem 
by adapting a model trained on a \textit{source} domain to a \textit{target} domain. 
In particular, unsupervised DA (UDA) assumes that only unlabeled samples are available from the target domain, whereas labels are available for samples from the source domain. 
We present a new UDA method in this paper.

While a number of UDA approaches have been proposed, many recent advances conform to the theory proposed by Ben-David et al. \cite{Ben10}. 
Ben-David et al. formulated the domain shift in terms of domain divergence. 
More specifically, they demonstrated that the target error is bounded by the source error and the divergence between the two domains. 
Hence, to realize a successful adaptation of a source-trained model to the target domain, small source-target divergence is essential.

In this regard, the majority of modern DNN-based UDA methods \cite{DANN,Tzeng15,CoGAN,ADDA} work by learning the features that minimize the domain divergence. 
To achieve this goal, they typically employ a domain discriminator that separates the source against the target samples. 
This domain discriminator empirically measures the domain divergence using unlabeled data from both domains. 
In these methods, a feature extractor is trained in an adversarial manner to maximize the confusion of the domain discriminator and to align the distributions of features between the source and the target domains. 
The underlying assumption is that once domain-invariant features are learned, a model trained on source data will correctly classify the target data.
However, these approaches show limited performance in reducing the target error because they inherently focus on matching marginal distributions from both domains. 
Though the domain discriminator is useful in learning domain-invariant features,
it lacks class-aware information \cite{Ben10,MCD}. 
A recent study \cite{ICML2019_Zhao} also suggests that the target error can increase when the label distributions of both domains differ significantly. 

\begin{figure*}[!t]
    \centering
    \includegraphics[width=0.95\textwidth]{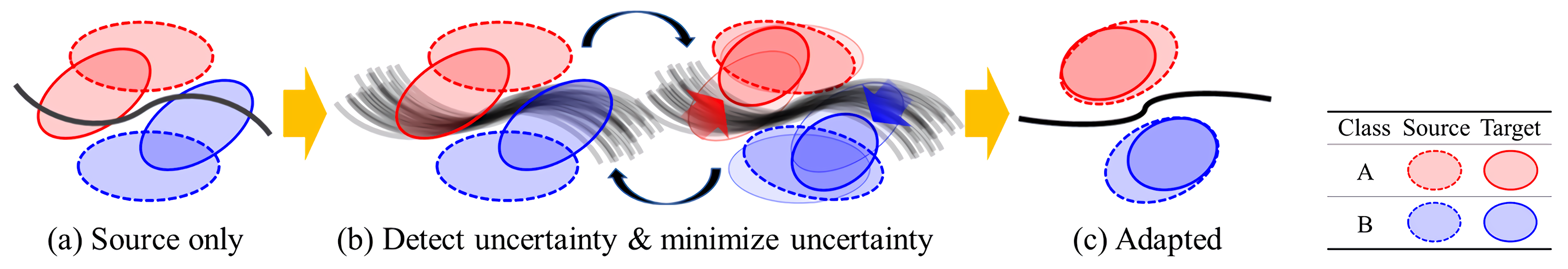}
    \caption{(Best viewed in color) Our approach exploits \textit{model uncertainty} to detect and minimize the divergence between hypotheses.}
    \label{fig:intro}
\end{figure*}

To resolve these issues, we take class-aware information into consideration for domain adaptation.
In this work, we propose predictive uncertainty as a domain divergence measure.
Our intuition is that less discriminative target samples that cause domain discrepancy are likely to show higher predictive uncertainty when inferred with source-only models.
Thus, if we can train a feature extractor that minimizes model uncertainty on target samples, it will generate features that are more consistent across different domains. 
This process is illustrated in Fig. \ref{fig:intro}.
We justify our approach by associating the model uncertainty with the classifier-induced domain divergence. 
According to \cite{Ben10}, this divergence provides a theoretically tighter bound than domain discriminator-based divergence.

A related study \cite{MCD} proposed the disagreement between two task-specific classifiers as a measure of domain divergence.
This method trains two classifiers on labeled source samples and computes their discrepancy with the $L_1$-distance between their softmax outputs for target samples. 
A feature generator is trained to minimize the maximum classifier discrepancy. 
A similar approach appears in \cite{GPDA}, where a Gaussian process model was proposed, with the maximum classifier discrepancy translated into the maximum posterior separation.
Variational inference is used to train a feature embedding network that maximizes the margin between top-1 and top-2 predictions.

However, these methods are overly conservative because domain divergence is measured with only two classifiers. 
A recent work \cite{STAR} reports that using more classifiers improves UDA performance.
Whereas two classifiers may fail to identify misaligned features, this failure is less likely with more classifiers. 
Therefore, we relax the restriction by considering the expectation over an approximately infinite number of classifiers. 
To this end, we adopt a Bayesian framework and define a distribution of classifiers given the source data.
We show that this modified domain divergence is related to our model uncertainty.
We also present Monte Carlo (MC) dropout sampling as an efficient way to assess this model uncertainty.  
Our approach is motivated by the recent interpretation of dropout as approximate Bayesian inference \cite{Gal16,Gal16iclr}. 
Thus, we can evaluate the model uncertainty from multiple feedforward passes on the target samples.

We evaluate our method on various image recognition tasks to demonstrate its effectiveness. 
Our contributions are summarized below.
\begin{itemize}
\item 
We develop a novel interpretation of the model uncertainty as a measure of the divergence between domains and propose a new UDA method based on the model uncertainty (MUDA).

\item 
We propose to use dropout as a training method under the Bayesian approximation scheme. 
Hence, the proposed MUDA can be applied to any DNN model that supports dropout. 

\item 
Our extensive experimental results demonstrate that MUDA achieves state-of-the-art outcomes on popular benchmark datasets. 
We also show that MUDA performs competitively on multi-source DA problems despite the simple source-combine strategy.

\item
We present several qualitative analyses to verify that the proposed method successfully captures discriminative features for the target tasks.

\end{itemize}

A preliminary version of this work appeared in \cite{MUDA}.
With respect to \cite{MUDA}, this work includes substantially extended experiments and analyses.

\section{Background}
\label{sec:background}

\subsection{Problem Statement}
\label{ssec:uda}
We are given fully-labeled data $(X_s,Y_s)$ from the source domain distribution 
$\D_S$ and unlabeled data $X_t$ from the target domain distribution $\D_T$.
The goal of UDA is to build a classifier that correctly predicts the label 
$y_t \in \{1, \ldots, K\}$ of a new target sample $x_t$. 
Both domains are assumed to share the same set of labels.

\subsection{Related Work}

\noindent \textbf{Ben-David Theorem}
\label{ssec:ben-david}
Ben-David et al. \cite{Ben10} proposed the theory that provides the upper-bound of the expected target error of a hypothesis $\epsilon_T(h)$, using three terms: (i) the expected source error $\epsilon_S(h)$; (ii) the $\H{\triangle}\H$ divergence between the two domains; and (iii) a hypothesis-independent constant $\lambda$, as follows:
\begin{equation}
  ^{\forall}h \in \H, ~~
  \epsilon_T(h) 
  \leq \epsilon_S(h) + \frac{1}{2}d_{\H\triangle\H}(\D_S, \D_T) + \lambda
  \label{eq:bd_thm}
\end{equation}
where $\H$ is a hypothesis space of classifiers and $d_{\H{\triangle}\H}$ denotes the divergence in the symmetric difference hypothesis space.
The constant $\lambda = \epsilon_S(h^*) + \epsilon_T(h^*)$ represents the combined error of the ideal joint hypothesis $h^*$ and is assumed to be small in UDA. 
This inequality suggests that minimizing the source error $\epsilon_S$ and divergence $d_{\H{\triangle}\H}$ is essential.

\medskip
\noindent \textbf{Domain Adaptation}
Based on the theory in \cite{Ben10}, a number of current UDA studies have focused on reducing the domain divergence. 
Earlier works \cite{MMD, DAN} aligned the feature statistics of the source and the target domains. 
More recent developments adopted the adversarial learning strategy to transfer knowledge between domains, inspired by GAN \cite{GAN}. 
These methods typically employ a domain discriminator to measure the difference between the two domains. 
DANN \cite{DANN} is the first to introduce this domain classifier.
To learn domain-invariant features, DANN trains its model with gradient-reversed backpropagation. 
ADDA \cite{ADDA} makes two domain-specific feature extractors which are trained to generate embeddings that confuse the domain discriminator. 
A large family of works, including JAN \cite{JAN}, MADA \cite{MADA}, and CADA \cite{CADA}, also train their models in an adversarial fashion. 
Saito et al. employ two instances of task-specific classifiers to use as a domain discriminator \cite{MCD, ADR}. 
Their method, MCDA, measures the domain divergence by the maximum discrepancy between two classifiers on the target samples. 
MCDA trains a feature generator to push the target feature distribution away from the decision boundary. 
GPDA(2019) \cite{GPDA} introduces the Gaussian Process (GP) model to extend earlier results \cite{MCD} and reformulates the maximum classifier discrepancy principle into the maximum posterior separation.  
STAR \cite{STAR} extends MCDA by modeling classifier weights with a Gaussian distribution where its variance represents the inter-classifier discrepancy. 
In other lines of approach, SAFN \cite{SAFN} argues that smaller feature norms cause model degradation and proposes to adapt the norms to a larger range.  
SWD \cite{SWD} uses the Wasserstein distance in domain alignment. 
Optimal transport strategy is used in DeepJDOT \cite{DeepJDOT} and RWOT \cite{RWOT}. 
DMRL \cite{DMRL} introduces mixup regularization to adversarial domain adaptation. 
GPDA(2021) \cite{GPDA21} performs graph dual regularization to preserve data statistics and geometric properties. 
DMAT \cite{DMAT} adversarially trains a dual-module network to learn domain invariant features and domain discriminative features separately.

\medskip
\noindent \textbf{Variational Inference}
In a Bayesian framework, a model $h$ such as the GP has a set of random parameters $\omega$ with a prior distribution $p(\omega)$.
Given a dataset $(X, Y)$, the predictive distribution for a new sample $x$ is given by 
\begin{gather}
  p(y| x, X, Y) = \int p(y| x, \omega) ~ p(\omega| X, Y) ~ d\omega
\end{gather}
where $p(\omega| X, Y)$ is the posterior distribution. 
However, exact evaluation of this posterior distribution is usually intractable. 
In variational inference, we approximate the posterior distribution 
$p(\omega| X, Y)$ with a simpler distribution $q(\omega)$. 
To make $q(\omega)$ similar to $p(\omega| X, Y)$, we minimize the Kullback-Leibler (KL) divergence $\text{KL}(q(\omega) || p(\omega| X, Y))$.
The KL divergence minimization is equivalently achieved by maximizing the evidence lower bound
\begin{gather}
  \int q(\omega) ~ \log p(Y| X, \omega) ~d\omega \
	- \text{KL}(q(\omega) || p(\omega))
\end{gather}
with respect to $q(\omega)$ \cite{PRML}.
By replacing the true posterior distribution $p(\omega| X, Y)$ with its variational approximation $q(\omega)$, we obtain the approximate predictive distribution
\begin{gather}
  q(y|x) = \int p(y| x, \omega) ~ q(\omega) ~ d\omega.
\end{gather}
Gal et al. \cite{Gal16} have recently shown that dropout in DNNs is mathematically identical to approximate variational inference in the deep GP.

\medskip
\noindent \textbf{Monte Carlo Dropout}
In our method, it is necessary to evaluate the predictive uncertainty of a model on the target samples.
Probabilistic Bayesian models can be used to estimate the model uncertainty, but these usually come with a prohibitive computational cost.
It was recently shown that dropout is equivalent to an approximation to a deep Bayesian model.
Whereas dropout is commonly used to prevent overfitting of a DNN model \cite{Dropout}, Gal et al. established a theoretical interpretation of dropout as approximate Bayesian inference to the deep GP model \cite{Gal16,Gal16iclr}.
As a result, dropout can be used to obtain the model uncertainty. 
This technique, referred to as MC dropout, enables to estimate the model uncertainty using an ensemble of multiple stochastic feedforward passes.

\section{Proposed Method}
\label{sec:method}

\subsection{Model Preliminaries}
\label{ssec:model_prelim}
We formulate the UDA problem similarly to recent UDA methods. 
A feature extractor network $F$ takes a sample $x$ and produces a latent feature vector $\phi = F(x)$. 
A task-specific classifier network $C$ takes $\phi$ and produces a $K$-dimensional output. 
$F$ and $C$ are shared by both the source and the target domains. 
Given $x$, the classification outcome is $\yhat = C(F(x))$, where $\yhat$ is a 
$K$-dimensional vector containing softmax scores.
The decision vector determined by $\yhat$ is denoted by $h(x)$.

\subsection{Unsupervised Domain Adaptation Based on Model Uncertainty}
\label{ssec:model_prelim}

A domain shift occurs when the source samples fail to represent the target samples. 
Because such target samples are less discriminative, their predictive uncertainty is likely to be high, as shown in Fig. \ref{fig:intro}b. 
High predictive uncertainty indicates that the target samples are outside of the source distribution. 
This leads to our idea that if we find a feature extractor $F$ that minimizes the predictive uncertainty, it will avoid generating target features outside of the source distribution.

The inequality \eqref{eq:bd_thm} implies that reducing the source error $\epsilon_S$ and the divergence $d_{\H \triangle \H}$ is essential for successful UDA.
We describe how we accomplish this goal by reducing the model uncertainty. 
The divergence $d_{\H \triangle \H}$ in \cite{Ben10} is defined as follows: 
\begin{align} \label{eq:symm_div}
  d_{\H \triangle \H}(\D_S, \D_T)
  = 2 \sup_{h, h' \in \H} 
    \big| & \E_{x \sim \D_S}[\ind(h(x) \neq h'(x))] 
		- \E_{x \sim \D_T}[\ind(h(x) \neq h'(x))] \big|.
\end{align}
If $h$ and $h'$ can correctly classify the source samples, we can consider that they will agree on the source samples. 
This enables us to safely neglect the term $\E_{x \sim \D_S}[\ind(h(x) \neq h'(x))]$.

Because $\ind(h(x) \neq h'(x)) = (h(x) - h'(x))^2$ for the binary classification 
$h(x) \in \{0, 1\}$, we can approximate $d_{\H \triangle \H}$ by 
\begin{align}
  d_{\H \triangle \H}(\D_S, \D_T)
  ~\approx~ & 2 \sup_{h,h'{\in}\H} \E_{x\sim \D_T}[\mathbbm{1}(h(x){\neq}h'(x))]
  \label{eq:app_hsym}
  \\
  =~ & 2 \sup_{h,h'{\in}\H} \E_{x \sim \D_T}[(h(x) - h'(x))^2].
  \label{eq:div}
\end{align}
Hence, the UDA is simplified to the problem of minimizing the supremum of the expected disagreement between two hypotheses on the target samples.
Though this objective involves a supremum over all hypothesis functions in $\H$, we argue that we can achieve the same goal by narrowing our attention to the set of hypotheses that minimizes $\epsilon_S$.

To this end, we follow the Bayesian approach and define the posterior distribution $\D_\H = p(h | X_s, Y_s)$ of $h$ conditioned on the labeled source samples. 
We also manipulate equation \eqref{eq:div} by replacing the supremum with the expectation with respect to the posterior $\D_\H$ to obtain 
(see Appendix) 
\begin{align}
  & 2 ~\E_{h,h'\sim \D_\H} ~\E_{x \sim \D_T}[(h(x) - h'(x))^2] 
  \label{eq:before}
  \\
  =~ & 4 ~\E_{x \sim \D_T} ~\E_{h \sim \D_\H} [(h(x) - \E_{h \sim \D_\H} [h(x)])^2]. 
  \label{eq:after}
\end{align}
Therefore, we can reformulate the problem of minimizing the divergence $d_{\H \triangle \H}$ into minimizing the predictive variance of a hypothesis ({\it model uncertainty}) on target samples.
This replacement will no longer ensure the modified divergence in equation \eqref{eq:after} as an upper bound of the target error.
However, we can easily imagine that minimizing the mean hypothesis disagreement will lead to similar consequences as minimizing the supremum hypothesis disagreement.
In section \ref{ssec:discussion}, we provide more discussion of this reformulation. 
Consequently, our objective is to find a feature extractor $F$ that minimizes this model uncertainty:
\begin{gather}
  \min_F ~\E_{x \sim \D_T} ~\E_{h \sim \D_\H} [(h(x) - \E_{h \sim \D_\H} [h(x)])^2] 
  \label{eq:obj_div}
\end{gather}
while minimizing the source error $\epsilon_S$:
\begin{gather}
  \min_{F, C} ~\epsilon_S (h).
  \label{eq:obj_cls}
\end{gather}
We note that the extension to $K$-way classification is straightforward, where $h(x)$ is a $K$-dimensional decision vector.

\subsection{Model Uncertainty Loss}
\label{ssec:uncertainty_loss}

Though we can model the posterior $\D_\H = p(h|X_s, Y_s)$ with a GP, its evaluation is often intractable. 
Variational inference instead defines an approximate posterior $q(\omega)$ and yields an approximate predictive distribution for a new target sample $x \sim \D_T$, as follows:
\begin{gather}
  q(y|x) = \int p(y|x, \omega) ~ q(\omega) ~ d\omega
\end{gather}
where $\omega$ is a set of random parameters for $h$.

Since dropout applied to a DNN is shown to be equivalent to the approximate variational inference of the deep GP \cite{Gal16}, we can evaluate the predictive variance in equation \eqref{eq:obj_div} by means of MC dropout sampling.
By performing $M$ stochastic forward passes through the DNN for $h$ (equivalently, $F$ and/or $C$), we obtain $\{ \yhat_1 \ldots \yhat_M \}$, where $\yhat_m$ is a $K$-dimensional softmax score.
These computations can be done in parallel.
Our estimate of the predictive variance then becomes 
\begin{gather}
  \widehat{\sigma}_{\MC}^2 (x)
  = \text{diag}(\frac{1}{M} \sum_{m=1}^{M} \yhat_m{\yhat_m}^\top 
    - \overline{y}_{\MC} \overline{y}_{\MC}^\top),
  \label{eq:sigma}
  \\
  \text{where }~~
  \overline{y}_{\MC} = \frac{1}{M} \sum_{m=1}^{M} \yhat_m.
\end{gather}
Therefore, we define our model uncertainty loss with the norm of this predictive variance: 
\begin{gather}
  \L_{div}(\D_T) 
  = \E_{x \sim \D_T} \| \widehat{{\sigma}}_{\MC} (x) \|
  \label{eq:loss_div}
\end{gather}
as an approximation to the divergence $d_{\H \triangle \H}$.

\begin{algorithm}[!t]
\caption{Learning algorithm for MUDA}
\label{alg2:muda}
\textbf{Input:} Labeled source data $(X_s,Y_s)$; unlabeled target data $X_t$; 

\hspace{11mm} MC dropout sample size $M$; MC dropout rate ($\rho_F$, $\rho_C$).

\textbf{Output:} Domain-adapted feature extractor $F$ and classifier $C$.
\begin{algorithmic}[1]
\State Learn initial weights of $F$ and $C$ on $(X_s,Y_s)$.

        $\qquad \quad 
          F, C \leftarrow \arg\min_{F, C} \L_{cls}(\D_S)$

\Repeat

  \State Sample mini-batch $(X_s, Y_s)^{\text{mini}}$ from $(X_s,Y_s)$.
          
  \State Sample mini-batch $X_t^{\text{mini}}$ from $X_t$.
  
  \State Compute $\L_{cls}(\D_S)$ on $(X_s, Y_s)^{\text{mini}}$ 
          by equation \eqref{eq:loss_cls}.
  
  \State {\it Activate} dropout of $F$ and $C$ with ($\rho_F$, $\rho_C$).
  
  \For{$m=1$ to $M$}
    \State Perform stochastic feedforward on $X_t^{\text{mini}}$.
      
        $\qquad \quad 
          \widehat{Y}_m^{\text{mini}} \leftarrow C(F(X_t^{\text{mini}}))$
  \EndFor

  \State {\it Deactivate} dropout of $F$ and $C$.
  
  \State Compute $\widehat{\sigma}_{\MC}^2$ using 
      $\{ \widehat{Y}_1^{\text{mini}}, \ldots, \widehat{Y}_M^{\text{mini}} \}$ 
      by equation \eqref{eq:sigma}.
  
  \State Compute $\L_{div}(\D_T)$ on $X_t^{\text{mini}}$ 
      by equation \eqref{eq:loss_div}.
  
  
  \State Update $C$:
        $\quad
          C \leftarrow \arg\min_C \L_{cls}(\D_S)$
  
  \State Update $F$:
        $\quad
          F \leftarrow \arg\min_F \L_{cls}(\D_S) + \L_{div}(\D_T)$
  
\Until{Stopping conditions are satisfied}
\end{algorithmic}
\end{algorithm}

\subsection{Classification Loss}
\label{ssec:classification_loss}
Keeping the source error $\epsilon_S(h)$ minimized is essential for successful UDA.
To measure the source error, we use the cross entropy loss 
\begin{gather}
  \L_{cls} (\D_S) = - \E_{(x_s, y_s) \sim \D_S} [y_s^{\top} \log C(F(x_s))]
  \label{eq:loss_cls}
\end{gather}
for our $K$-way classification task. 
Here, $y_s$ denotes the one-hot encoded label vector of a source sample $x_s$.

We note that this task-specific loss can be replaced according to the purpose of the given task.
For example, one might use the mean squared loss for a regression task or the pixel-wise cross entropy loss for a semantic segmentation task.

\subsection{Optimization Strategy}
\label{ssec:opt_strategy}

As described above, our $d_{\H \triangle \H}$-divergence approximation and the source error in equations \eqref{eq:obj_div} and \eqref{eq:obj_cls} can be evaluated with the model uncertainty loss $\L_{div}(\D_T)$ and the cross entropy loss $\L_{cls}(\D_S)$, respectively. 
This leads to two optimization problems:
\begin{align}
  & \bullet  \quad  \min_C ~ \L_{cls}(\D_S) \label{opt:cls} \\
  & \bullet  \quad  \min_F ~ \L_{cls}(\D_S) + \L_{div}(\D_T).
  \label{opt:all}
\end{align}
Therefore, our domain adaptation method based on model uncertainty (MUDA) alternates to train $F$ and $C$. 
The proposed method is outlined in Algorithm 1.

\begin{figure}[!t] 
\centering
  \includegraphics[width=0.8\columnwidth]{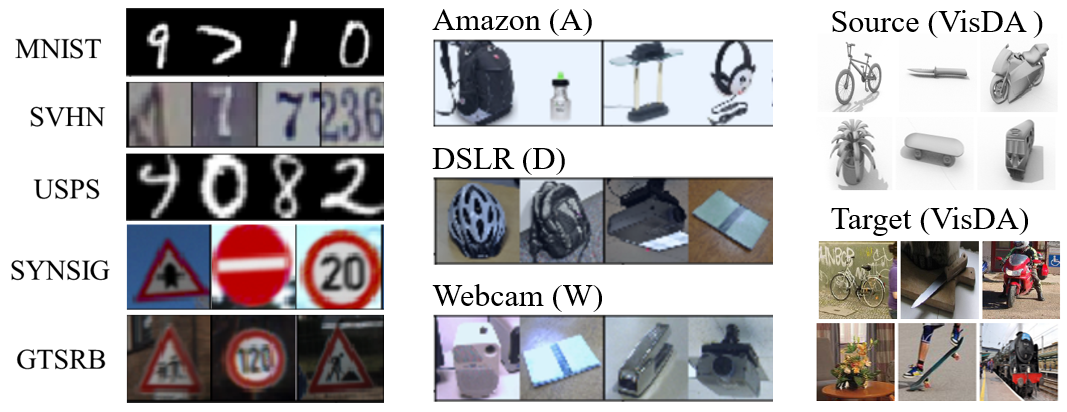}
    \caption{Sample images of domain adaptation benchmark datasets.}
    \label{fig:uda_dataset}
\end{figure}

\section{Experiments}
\label{sec:exp}

To demonstrate our approach, we conduct experiments on various datasets. 
A proof-of-concept experiment on a toy dataset (section \ref{ssec:toy_dataset}) 
is followed by an extensive evaluation on the digits and traffic signs 
(section \ref{ssec:digit_dataset}), the Office-31 (section 
\ref{ssec:office31_dataset}), and the VisDA-17 (section 
\ref{ssec:visda17_dataset}) datasets.
Fig. \ref{fig:uda_dataset} shows a few sample images from each of these 
benchmark datasets.

We extend our UDA method to problem in which more than one source domain datasets are available. 
Digits-five is a collection of digit images from five different domains, and miniDomainNet \cite{DAEL} contains object images from four different domains.
MUDA shows competitive results compared to recent multi-source domain adaptation methods (section \ref{ssec:further}). 

We also qualitatively analyze our method by visualizing the changes of feature embeddings and by comparing class activation maps with or without MUDA 
(section \ref{ssec:discussion}).
All experiments are implemented using PyTorch \cite{PyTorch}.
Our code is available at \url{https://github.com/joonholee-research/MUDA}.

\subsection{Proof of Concept}
\label{ssec:toy_dataset}

\begin{figure}[!t]
  \centering
  \begin{minipage}[t]{0.32\linewidth} \centering \footnotesize
  \includegraphics[width=\linewidth]{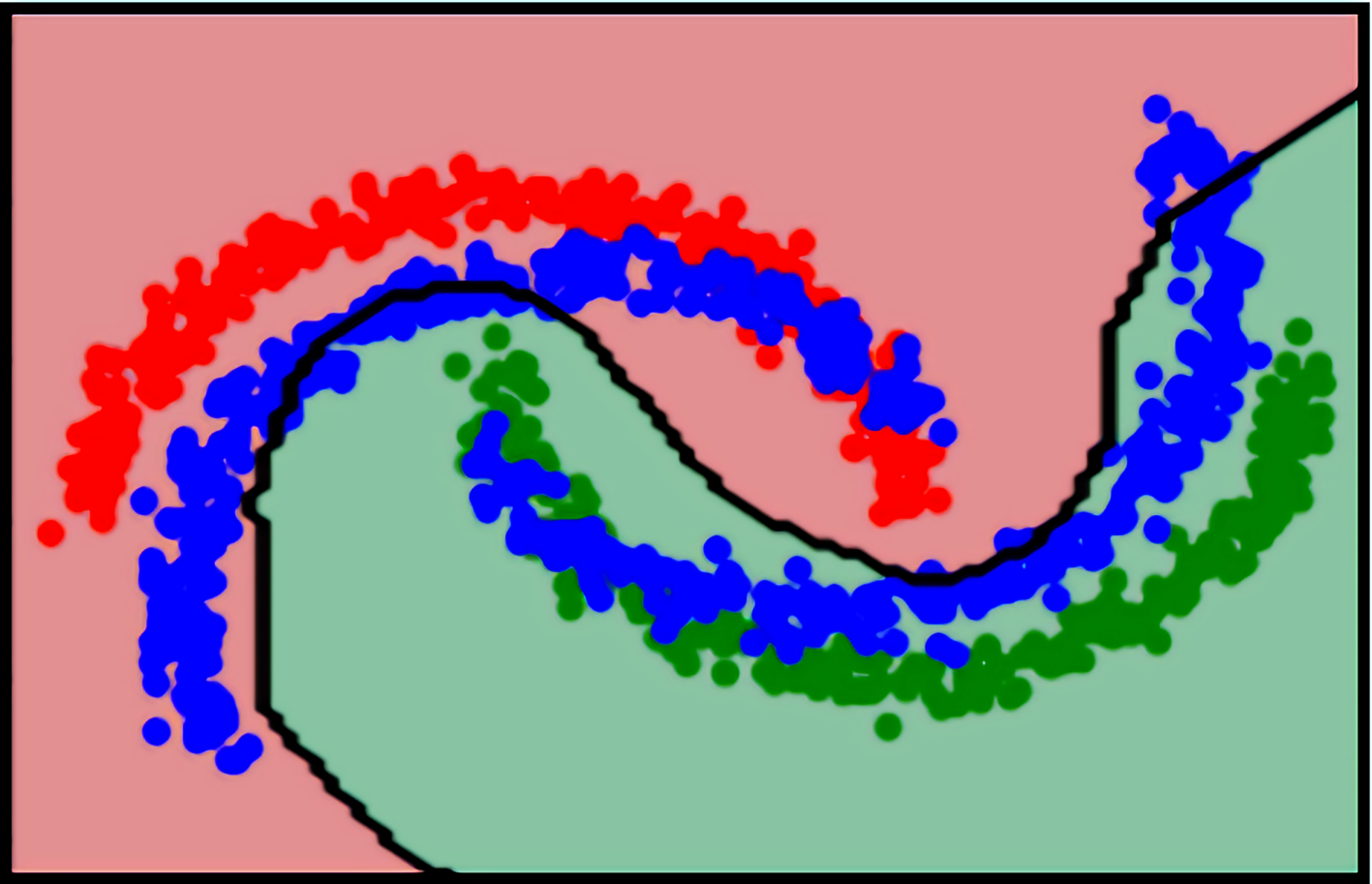}
  (a) Source-only
  \end{minipage}
  \begin{minipage}[t]{0.32\linewidth} \centering \footnotesize
  \includegraphics[width=\linewidth]{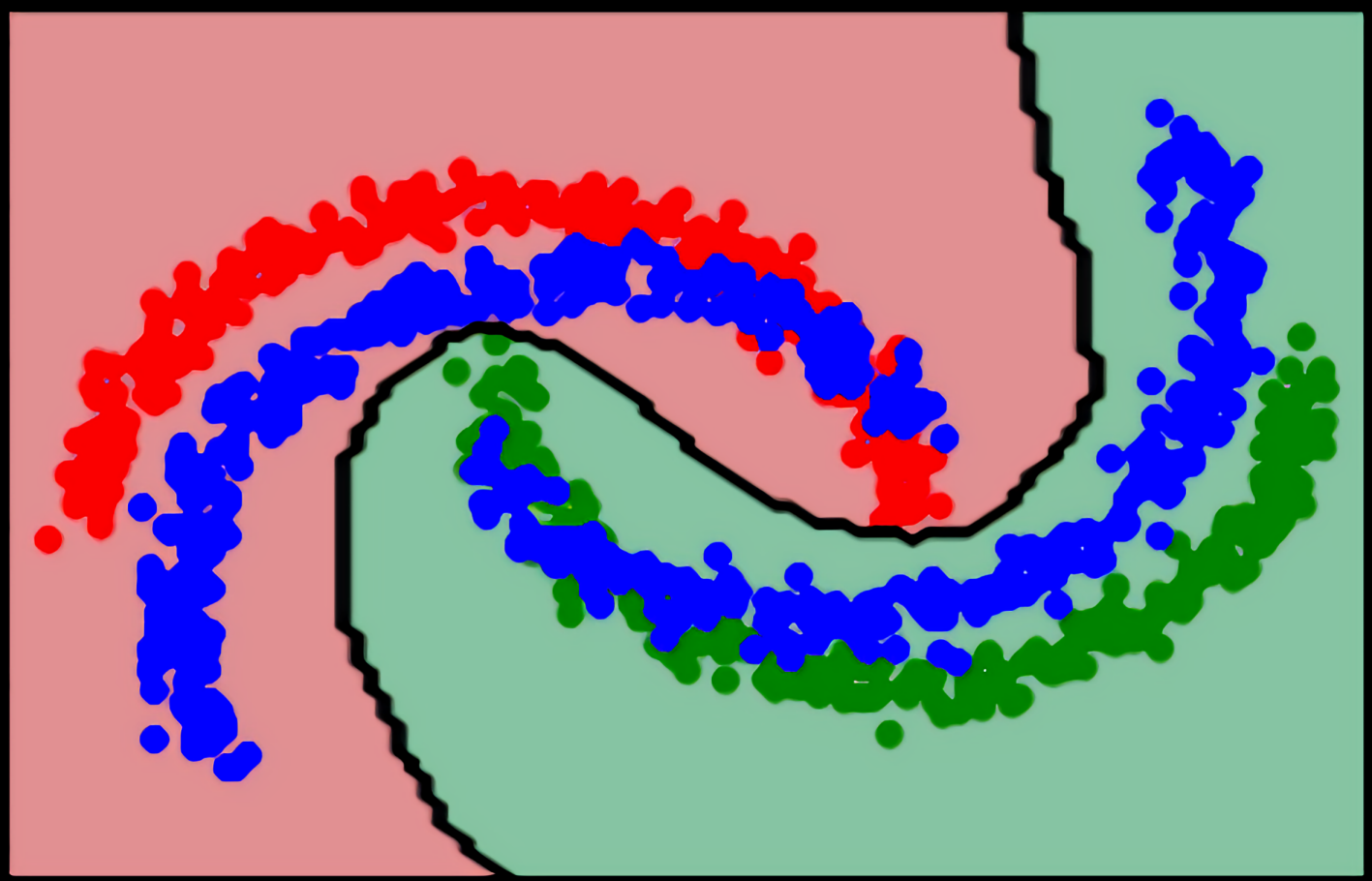}
  (b) MUDA
  \end{minipage}
  \begin{minipage}[t]{0.32\linewidth} \centering \footnotesize
  \includegraphics[width=\linewidth]{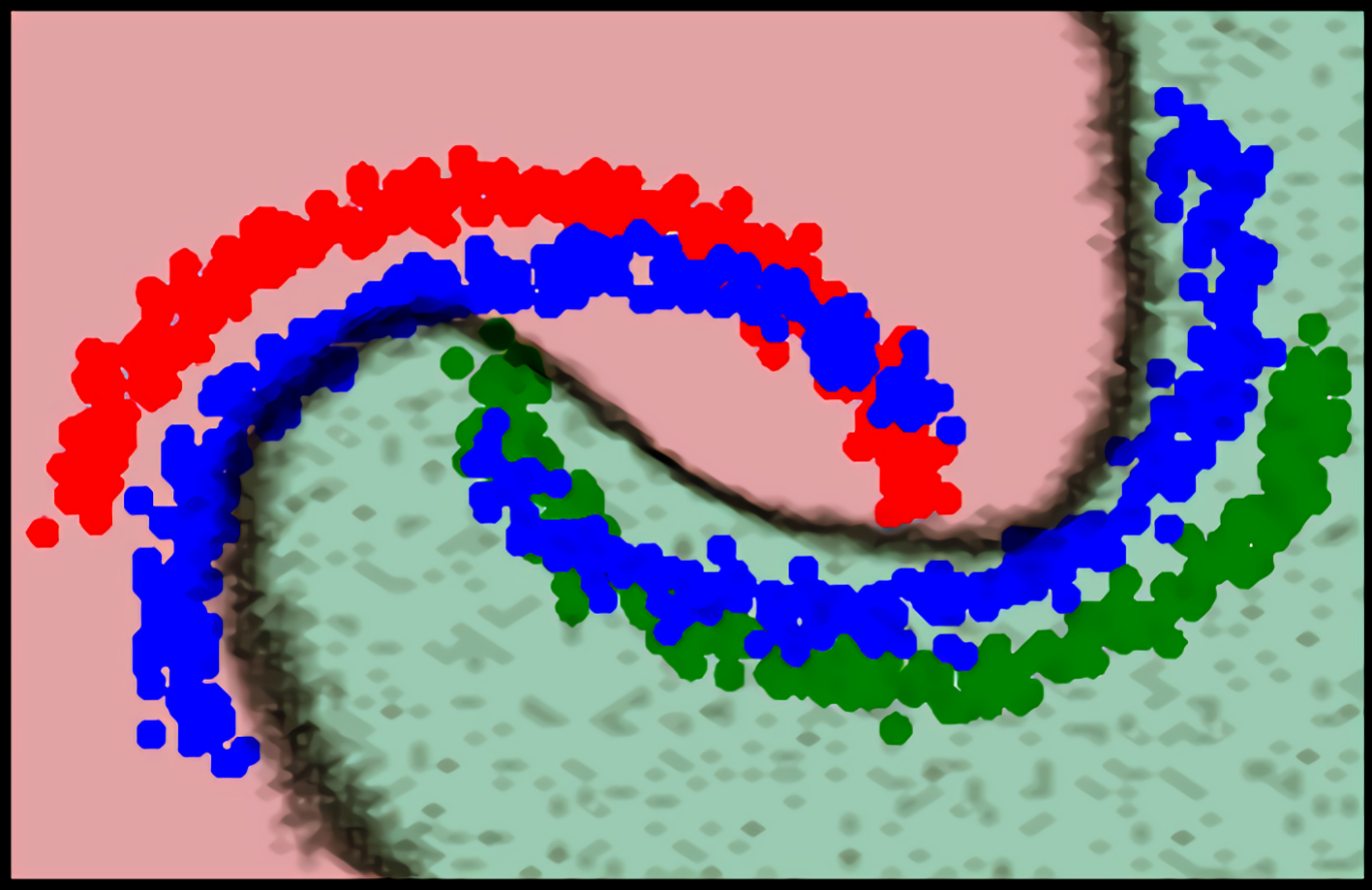}
  (c) MUDA(MC dropout)
  \end{minipage}
  \caption{(Best viewed in color) Red and green dots are color-labeled source samples, and blue dots represent unlabeled target samples. The black curve represents a decision boundary. The pink and light green areas are the decision regions of the classifier. (a) shows the domain shift. The adapted decision boundary in (b) correctly classifies target samples.
  (c) displays multiple boundaries from MC dropout in our MUDA method.
  Note that accumulated boundary is thicker near the target samples far from the source distribution. }
  \label{fig:toy}
\end{figure}

\noindent {\bf Setup}
We conduct a proof-of-concept experiment on the interleaving two-moons dataset. 
Fig. \ref{fig:toy} shows the labeled source samples (red, green) and the unlabeled target samples (blue).
The target distribution is obtained by rotating the source distribution by $30^\circ$. 
For the experiment, we generate 1000 source and target samples, respectively, and split them into 500 training samples and 500 testing samples for each domain. 
The test samples from both domains are depicted in Fig. \ref{fig:toy}.

We use a five-layer fully-connected (FC) network with 15 hidden neurons for every hidden layer. 
Each of the first three layers is followed by batch normalization, and the fourth layer is set to dropout at a rate 0.5. ReLU is used for activation. The mini-batch size is 128.

\medskip
\noindent {\bf Results}
Fig. \ref{fig:toy} compares the model adapted using our MUDA method with a model trained only on source samples.
The decision boundary of the source-only model in Fig. \ref{fig:toy}a crosses the support of the target distribution and misclassifies many target samples. 
On the other hand, the proposed MUDA adapts the model to the target samples and successfully separates them with high accuracy as shown in Fig. \ref{fig:toy}b. 
MUDA achieves this by finding the decision boundary that minimizes the predictive uncertainty in the target domain. 
Note that the decision boundary is moved away from the target samples. 
Fig. \ref{fig:toy}c illustrates how MUDA measures the predictive uncertainty. 
Twenty classifiers from MC dropout sampling are overlaid. 
The accumulated boundary is thicker where the boundary is closer to the target samples and farther from the source samples, indicating that the predictive uncertainty is higher.

\subsection{Digits and Traffic Signs Datasets}
\label{ssec:digit_dataset}

\noindent {\bf Setup}
We compare the proposed method with recent UDA methods in the literature on standard benchmark datasets. 
The digit classification task consists of three datasets: MNIST \cite{MNIST}, SVHN \cite{SVHN}, and USPS \cite{USPS}. 
Its aim is to classify an image into one of ten digit classes. 
We also conduct an evaluation with a traffic sign classification task. 
SYNSIG \cite{SYNSIG} and GTSRB \cite{GTSRB} contain 43 types of synthetic and actual signs, respectively.

We utilize the experimental setup with the dataset splits and the network architectures of prior works \cite{DANN, ADDA, MCD} for a fair comparison.
For MNIST$^*$ $\rightarrow$ USPS$^*$, we use all of the target samples during training. 
For all other tasks, we split both the source and the target samples into the training and test datasets. 

Three convolution layers for $F$ and two FC layers for $C$ are used with intermediate dropout layers. 
The input image size is set to 28$\times$28 for MNIST $\leftrightarrow$ USPS, 32$\times$32 for SVHN $\rightarrow$ MNIST, and 40$\times$40 for SYNSIG $\rightarrow$ GTSRB. 
The Adam \cite{Adam} optimizer is used with a learning rate 2.0$\times$10$^{-4}$ and weight decay 5.0$\times$10$^{-4}$. 
The dropout rate is set to 0.4 for $C$ and 0.1 for $F$. 
The mini-batch size is 64.

\medskip
\noindent {\bf Results}
Table \ref{tab:digits} summarizes the results on the digits and traffic signs 
datasets. 
The mean and the standard deviation of ten independent experiments are reported. 
The table also compares recent state-of-the-art UDA methods. 
As can be seen, MUDA exhibits superior performance for all tasks. 
The improvement over the source-only model ranges from 13.5\% to 33.3\%.
In particular on a difficult task, SVHN $\rightarrow$ MNIST, MUDA increases the accuracy by more than 30\%. 
Whereas no labels are used, the performance is nearly perfect.
Though some methods like STAR\cite{STAR} and RWOT\cite{RWOT} show slightly higher accuracies for USPS $\rightarrow$ MNIST, the differences are not significant.
\begin{table}[!t]
\centering
\caption{Classification accuracy (\%) on the digits and traffic signs datasets 
(S: SVHN, M: MNIST, U: USPS, S $\shortrightarrow$ G: SYNSIG $\rightarrow$ GTSRB).
The results are cited from each study. We indicate the best in \textbf{bold} and the second best in \textbf{\textit{bold italic}}.} 

\smallskip
\def\arraystretch{1.2}
\resizebox{0.65\textwidth}{!}
{%
\begin{tabular}{l|c c c c c}
\Xhline{2\arrayrulewidth}
Methods 
& S $\shortrightarrow$ M
& M $\shortrightarrow$ U
& M$^*$ $\shortrightarrow$ U$^*$
& U $\shortrightarrow$ M
& S $\shortrightarrow$ G \\
\Xhline{2\arrayrulewidth}
Source-only & 67.1 & 76.7 & 79.4 & 63.4 & 85.1 \\ 
DAN)\cite{DAN} & 71.1 & - & 81.1 & - & 91.1\\
DANN\cite{DANN} & 71.1 & 77.1$^{ 1.8}$ & 85.1 & 73.0$^{ 0.2}$ & 88.7 \\
DSN\cite{DSN} & 82.7 & 91.3 & - & - & 93.1 \\
ADDA\cite{ADDA} & 76.0$^{ 1.8}$ & 89.4$^{ 0.2}$ & - & 90.1$^{ 0.8}$ & - \\
G2A\cite{G2A} & 92.4 & 92.8 & 95.3 & 90.8 & - \\
MCDA\cite{MCD} & 96.2$^{ 0.4}$ & 94.2$^{ 0.7}$ & 96.5$^{ 0.3}$ & 94.1$^{ 0.3}$ & 94.4$^{ 0.3}$ \\ 
GPDA(2019)\cite{GPDA} & {98.2}$^{ 0.1}$ & \textbf{\textit{96.5}}$^{ 0.2}$ & {98.1}$^{ 0.1}$ & 96.4$^{ 0.1}$ & {96.2}$^{ 0.2}$ \\
CADA\cite{CADA} & 90.9$^{ 0.2}$ & 96.4$^{ 0.1}$ & - & {97.0}$^{ 0.1}$ & - \\
SWD\cite{SWD} & \textbf{\textit{98.9}}$^{ 0.1}$& -& 98.1$^{ 0.1}$& 97.1$^{ 0.1}$& \textbf{98.6}$^{ 0.3}$\\
STAR\cite{STAR} & 98.8$^{ 0.1}$& -& 97.8$^{ 0.1}$& {\bf 97.7}$^{ 0.1}$& 95.8$^{ 0.2}$\\
RWOT\cite{RWOT} & 98.8$^{ 0.1}$& -& \textbf{98.5}$^{ 0.2}$& \textbf{\textit{97.5}}$^{ 0.2}$& -\\
GPDA(2021)\cite{GPDA21} & - & 83.2 & - & 74.1 & - \\
DMAT\cite{DMAT} & \textbf{\textit{98.9}}$^{0.1}$ & - & 95.1$^{0.4}$ & 96.1$^{0.2}$ & 91.1$^{0.2}$\\
\hline
\textbf{MUDA} (ours) & \textbf{99.1}$^{ 0.4}$ & \textbf{97.9}$^{ 0.2}$ & \textbf{98.5}$^{ 0.1}$ & {96.7}$^{ 0.4}$ & \textbf{98.6}$^{ 0.5}$ \\
\Xhline{2\arrayrulewidth}
\end{tabular}
}%
\label{tab:digits}
\end{table}

\subsection{Office-31 Dataset}
\label{ssec:office31_dataset}

\noindent {\bf Setup}
The Office-31 dataset \cite{Office31} contains 4,652 images across 31 categories collected from three different domains: \textit{Amazon} (A), \textit{Webcam} (W), and \textit{DSLR} (D). 
Whereas each image is larger, the dataset is smaller than the digits and traffic signs datasets.
Because the Office-31 dataset is relatively small, we evaluate MUDA fully transductively as in previous works \cite{JAN, ADDA, GPDA}. 
We use all labeled source samples and all unlabeled target samples for training. 

We employ the pre-trained ResNet-50 \cite{ResNet} as our feature extractor $F$ and FC layers with 1,000 neurons as the classifier $C$.
Every image is resized to 256$\times$256, randomly flipped, and then cropped on center to 224$\times$224. 
Images are standardized using ImageNet \cite{ImageNet} statistics before they are fed into the DNNs. 
We optimize using Adam with a learning rate 2.0$\times$10$^{-5}$, weight decay 5.0$\times$10$^{-4}$, and mini-batch size 32. The dropout is applied only to $C$ at a fixed rate of 0.4.

\begin{table}[!t]
\centering
\caption{ Classification accuracy (\%) on Office-31 (A: Amazon, W: Webcam, D: DSLR). 
The results are cited from each study. \textbf{Bold} is the best and \textbf{\textit{bold italic}} is the second best.
}

\smallskip
\def\arraystretch{1.2}
\resizebox{0.80\textwidth}{!}
{%
\begin{tabular}{l|c c c c c c|c}
\Xhline{2\arrayrulewidth}
Methods 
& A $\shortrightarrow$ W
& D $\shortrightarrow$ W
& W $\shortrightarrow$ D
& A $\shortrightarrow$ D
& D $\shortrightarrow$ A
& W $\shortrightarrow$ A
& Average \\

\Xhline{2\arrayrulewidth}
Source-only & 68.4 & 96.7 & 99.3 & 68.9 & 62.5 & 60.7 & 76.1 \\ 
DANN\cite{DANN} & 82.0 & 96.9 & 99.1 & 79.7 & 68.2 & 67.4 & 82.2  \\ 
ADDA\cite{ADDA} & 86.2 & 96.2 & 98.4 & 77.8 & 69.5 & {68.9} & 82.8 \\ 
JAN\cite{JAN} & 85.4 & {97.4} & \textbf{\textit{99.8}} & 84.7 & 68.6 & \textbf{\textit{70.0}} & 84.3 \\ 
MADA\cite{MADA} & \textbf{90.0} & {97.4} & 99.6 & \textbf{\textit{87.8}} & {70.3} & 66.4 & {85.3} \\ 
GPDA(2019)\cite{GPDA} & 83.9 & 97.3 & \textbf{100.0} & 85.5 & \textbf{72.3} & 68.8 & 84.6 \\ 
SAFN\cite{SAFN} & \textbf{\textit{88.8}} & \textbf{\textit{98.4}} & \textbf{\textit{99.8}} & 87.7 & 69.8 & 69.7 & \textbf{\textit{85.7}} \\ 
GPDA(2021)\cite{GPDA21} &87.4 &98.4 &99.4 &85.8 &70.6 &{\bf 72.8} &\textbf{\textit{85.7}} \\
\hline
\textbf{MUDA} (ours) & {88.2} & \textbf{98.7} & \textbf{\textit{99.8}} & \textbf{90.0} & \textbf{\textit{71.2}} & {69.0} & \textbf{86.1} \\ 
\Xhline{2\arrayrulewidth}
\end{tabular}
}%

\vspace{-2ex}
\label{table:office31}
\end{table}

\medskip
\noindent {\bf Results}
The results are presented in Table \ref{table:office31}.
We find that MUDA outperforms on this dataset as well.
MUDA achieves the best or second best on most UDA tasks and records the highest overall average accuracy. 
Compared to the source-only model, the average accuracy across all tasks increases from 76.1\% to 86.1\%.
For certain tasks such as \textit{Amazon} $\rightarrow$ \textit{DSLR} and \textit{Amazon} $\rightarrow$ \textit{Webcam}, the accuracy increases significantly by nearly 20\%. 
Even when the number of source instances is not sufficiently large, as in \textit{DSLR} $\rightarrow$ \textit{Amazon} and \textit{Webcam} $\rightarrow$ \textit{Amazon}, the improvement exceeds 8\%.
Because \textit{Amazon} is known to be substantially different from the other two domains \cite{Office31}, this performance gain indicates that MUDA is effective on challenging real-world adaptation tasks.

\subsection{VisDA-17 Dataset}
\label{ssec:visda17_dataset}

\noindent {\bf Setup}
The VisDA dataset \cite{Visda17} is designed to evaluate an adaptation from synthetic-object to real-object images. 
The source images are 3D object models rendered under different angles and different lighting conditions. 
This dataset contains 152,397 synthetic images across 12 categories. 
The target domain images are actual objects in the same categories. 
They are collected from MSCOCO \cite{MSCOCO} and number 55,388 in total.
The image samples are displayed in Fig. \ref{fig:uda_dataset}.
In our experiment, each image is randomly cropped and resized to 224$\times$224 before being randomly flipped.

We use the pre-trained ResNet-101 \cite{ResNet} for the feature extractor $F$. 
For the three FC layers of the classifier $C$, we set the number of hidden neurons to 1,000. 
We also incorporate the class balance loss into the objective functions 
to account for the class imbalance of the VisDA dataset.
The configured setup is similar to \cite{MCD, GPDA} for a fair comparison.
For optimization, we use SGD with a learning rate 10$^{-4}$, weight decay 5.0$\times$10$^{-4}$ and momentum 0.9. 
The mini-batch size is 32. 
The dropout rate of the classifier network $C$ is set to 0.4.

\begin{table*}[!t]
\centering
\caption{Classification accuracy per category (\%) on VisDA-17. 
The results are cited from each study. The mean is reported from 5 random runs, and each run updates for 5 epochs. We indicate the best in \textbf{bold} and the second best in \textbf{\textit{bold italic}}.}
\smallskip
\def\arraystretch{1.2}
\resizebox{1.00\textwidth}{!}
{%
\begin{tabular}{l| c c c c c c c c c c c c| c}
\Xhline{2\arrayrulewidth}
Methods & plane & bcycl & bus & car & horse & knife & mcycl & person & plant & sktbrd & train & truck & Average \\ 
\Xhline{2\arrayrulewidth}
Source-only & 55.1 & 53.3 & 61.9 & 59.1 & 80.6 & 17.9 & 79.7 & 31.2 & 81.0 & 26.5 & 73.5 & 8.5 & 52.4 \\ 
DAN\cite{DAN} & {87.1} & 63.0 & 76.5 & 42.0 & {90.3} & 42.9 & {85.9} & 53.1 & 49.7 & 36.3 & {85.8} & 20.7 & 61.1    \\ 
DANN\cite{DANN} & 81.9 & {77.7} & {82.8} & 44.3 & 81.2 & 29.5 & 65.1 & 28.6 & 51.9 & 54.6 & 82.8 & 7.8 & 57.4  \\
MCDA\cite{MCD} & 87.0 & 60.9 & \textbf{\textit{83.7}} & {64.0} & 88.9 & \textbf{79.6} & 84.7 & {76.9} & {88.6} & 40.3 & 83.0 & 25.8 & 71.9 \\ 
DeepJDOT\cite{DeepJDOT} & 85.4& 73.4& 77.3& \textbf{87.3}& 84.1& 64.7& \textbf{\textit{91.5}}& \textbf{\textit{79.3}}& \textbf{\textit{91.9}}& 44.4& 88.5& \textbf{61.8}& \textbf{\textit{77.4}}\\ 
GPDA(2019)\cite{GPDA} & 83.0 & {74.3} & 80.4 & {66.0} & 87.6 & {75.3} & 83.8 & {73.1} & {90.1} & {57.3} & 80.2 & \textbf{\textit{37.9}} & 73.3  \\ 
SAFN\cite{SAFN} & \textbf{93.6}& 61.3& \textbf{84.1}& \textbf{\textit{70.6}}& \textbf{94.1}& \textbf{\textit{79.0}}& \textbf{91.8}& 79.6& 89.9& 55.6& \textbf{89.0}& 24.4& 76.1\\ 
SWD\cite{SWD} & 90.8& \textbf{82.5}& 81.7& 70.5& 91.7& 69.5& 86.3& 77.5& 87.4& \textbf{63.6}& 85.6& 29.2& 76.4\\ 
DMRL\cite{DMRL} &-&-&-&-&-&-&-&-&-&-&-&-&75.5\\
DMAT+2M\cite{DMAT} & 86.0 & 61.5 & 88.3 & 61.6 & 83.8 & 6.7 & 92.9 & 56.8 & 89.9 & 68.8 & 87.3 & 23.0 & 69.2  \\
\hline
\textbf{MUDA} (ours) & \textbf{\textit{92.2}} & \textbf{\textit{79.5}} & 80.8 & {70.2} & \textbf{\textit{91.9}} & 78.5 & {90.8} & \textbf{81.9} & \textbf{93.0} & \textbf{\textit{62.5}} & \textbf{\textit{88.7}} & {31.9} & \textbf{78.5}  \\ 
\Xhline{2\arrayrulewidth}
\end{tabular}
}%
\label{table:visda17}
\end{table*}

\medskip
\noindent {\bf Results}
The adaptation results in Table \ref{table:visda17} clearly demonstrate that the proposed MUDA greatly outperforms other UDA methods. 
Our method records the highest average per-category accuracy. 
In addition, MUDA achieves the best or the second best outcome for seven out of 12 categories. 
Even for the most challenging categories such as {\it truck}, {\it knife}, {\it skateboard} and {\it person}, the improvements over the source-only model exceed from 20\% to 60\%. 

We find these results on the VisDA dataset encouraging. 
For many applications, adaptation from synthetic to real data can reduce the annotation costs considerably. 
However, the large domain difference makes the adaptation task challenging. 
In this regard, the performance improvements on the VisDA dataset along with the previous SYNSIG $\rightarrow$ GTSRB task provide convincing evidence that MUDA can overcome such difficulties.

\subsection{Multi-Source Domain Adaptation}
\label{ssec:further}

We extend MUDA to more complex multi-source domain adaptation problems. 
Unlike typical single-source UDA, multi-source DA considers adaptation from more than one source domains to one unlabeled target domain.
To apply MUDA to a multi-source problem, we simply adopt the \textit{source combine} strategy, in which all the source domains are combined to train a single model.

\begin{figure}[!t] 
\centering
  \centering
  \begin{minipage}[t]{0.25\linewidth} \centering \footnotesize
  \includegraphics[width=1.0\linewidth]{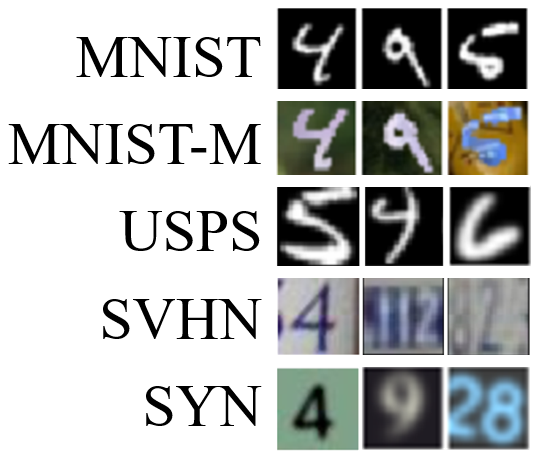}\\
  ~~~~~~(a) Digits-five 
  \end{minipage}~~~~~~~~~~~~
  \medskip
  \begin{minipage}[t]{0.30\linewidth} \centering \footnotesize
  \includegraphics[width=1.0\linewidth]{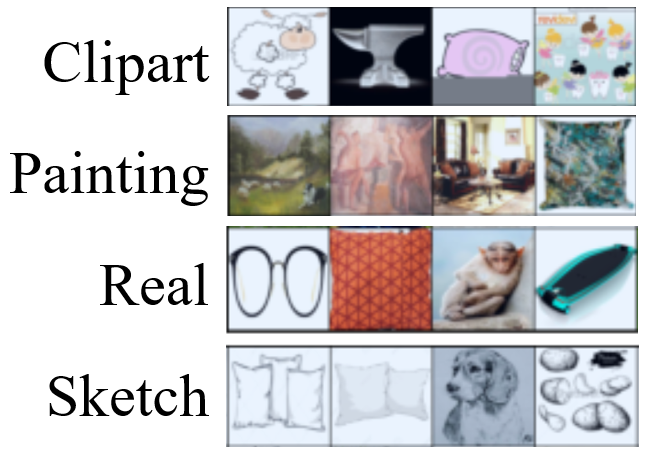}\\
  ~~~~~~(b) miniDomainNet 
  \end{minipage}
\caption{Sample images of Digits-five and miniDomainNet dataset.}
\label{fig:msda}
\end{figure}

\medskip
\noindent {\bf Setup}
We evaluate MUDA on Digits-five and miniDomainNet \cite{DAEL}.
Example images are illustrated in Fig. \ref{fig:msda}. 
Digits-five is a collection of digit images from five different domains: MNIST, MNIST-M \cite{DANN}, USPS, SVHN, and SYN \cite{DANN}. 
We follow \cite{M3SDA, DAEL} to sample 25,000 images for training and 9,000 for testing from each domain, except USPS. 
For the smaller USPS dataset, all 9,298 images are used. 
We choose one of the five domains for the target domain and the rest for the source domains. The experiment is repeated five times in turn.

MiniDomainNet \cite{DAEL} is a subset of DomainNet \cite{M3SDA} and contains 140,006 96$\times$96 images of 126 classes from four domains: Clipart, Painting, Real, and Sketch. 
From each domain of miniDomainNet, 630 test images are held out. 
Similarly, we repeat the experiment four times in turn.

For Digits-five, the same network architectures for $F$ and $C$ and the hyperparameters are used as in section \ref{ssec:digit_dataset}.
Each color image is randomly cropped and resized to 32$\times$32.
For miniDomainNet, ResNet-18 is used as in \cite{DAEL} and trained with the same hyperparameters in section \ref{ssec:office31_dataset}.

\begin{table*}[!t]
\centering
\caption{Classification accuracy (\%) on Digits-five multi-source experiment ({mt}: MNIST, {mm}: MNIST-M, {up}: USPS, {sv}: SVHN, {sy}: SYN, $\R$: the rest). 
The results are cited from each study. 
We indicate the best in \textbf{bold} in each standard. }
\smallskip
\def\arraystretch{1.2}
\resizebox{0.95\textwidth}{!}
{%
\begin{tabular}{c| c| c c c c c| c }
\Xhline{2\arrayrulewidth}
~~Standards~~&~~~~~Methods~~~~~ 
& $\R\rightarrow$ mt
& $\R\rightarrow$ mm
& $\R\rightarrow$ up
& $\R\rightarrow$ sv
& $\R\rightarrow$ sy
&~~~Average~~~\\ 
\Xhline{2\arrayrulewidth}
Oracle & Target-only & 99.65 & 98.30 & 99.52 & 94.30 & 99.22 & 98.20 \\ 
\hline
 & Source-only & 99.06 & 68.08 & 97.20 & 84.56 & 89.87 & 87.75 \\ 
Source & DANN \cite{DANN} & 98.46 $^{0.07}$ & 83.44 $^{0.12}$ & 94.19 $^{0.31}$ & 84.08 $^{0.60}$ & 92.91 $^{0.23}$ & 90.61 \\
Combine & MCDA \cite{MCD} & \textbf{99.22} $^{0.08}$ & 80.65 $^{0.51}$ & 98.32 $^{0.07}$ & 81.87 $^{0.72}$ & 95.42 $^{0.04}$ & 91.09 \\
 & \textbf{MUDA} (ours) & {99.18} $^{0.04}$ & \textbf{93.21} $^{0.25}$ & \textbf{98.89} $^{0.05}$ & \textbf{90.39} $^{0.29}$ & \textbf{97.02} $^{0.08}$ & \textbf{95.74}  \\ 
\Xhline{2\arrayrulewidth}
 & MDDA \cite{MDDA} & 98.80 & 78.60 & 93.90 & 79.30 & 89.70 & 88.06  \\ 
Multi-source & DCTN \cite{DCTN} & 99.38 $^{0.06}$ & 76.20 $^{0.51}$ & 94.39 $^{0.58}$ & 86.37 $^{0.54}$ & 86.78 $^{0.31}$ & 88.63  \\ 
UDA & M$^3$SDA \cite{M3SDA} & 99.38 $^{0.07}$ & 82.15 $^{0.49}$ & \textbf{98.71} $^{0.12}$ & 88.44 $^{0.72}$ & 96.10 $^{0.10}$ & 92.96  \\ 
 & DAEL \cite{DAEL} & \textbf{99.45} $^{0.02}$ & \textbf{93.77} $^{0.12}$ & 98.69 $^{0.79}$ & \textbf{92.50} $^{0.15}$ & \textbf{97.91} $^{0.03}$ & \textbf{96.47}  \\ 
\Xhline{2\arrayrulewidth}
\end{tabular}
}%
\label{table:msda}
\end{table*}
\begin{table*}[!t]
\centering
\caption{Classification accuracy (\%) on miniDomainNet (clp: Clipart, pnt: Painting, rel: Real, skt: Sketch, $\R$: the rest). 
The best in each standard is in \textbf{bold}. }
\smallskip
\def\arraystretch{1.2}
\resizebox{0.88\textwidth}{!}
{%
\begin{tabular}{c| c| c c c c | c }
\Xhline{2\arrayrulewidth}
~~Standards~~&~~~~~Methods~~~~~ 
& $\R\rightarrow$ clp
& $\R\rightarrow$ pnt
& $\R\rightarrow$ rel
& $\R\rightarrow$ skt
&~~~Average~~~\\ 
\Xhline{2\arrayrulewidth}
Oracle & Target-only & 72.59 & 60.53 & 80.47 & 63.44 & 69.26 \\ 
\hline
 & Source-only & 63.44 & 49.92 & 61.54 & 44.12 & 54.76 \\ 
Source & DANN \cite{DANN} & 65.55 $^{0.34}$ & 46.27 $^{0.71}$ & 58.68 $^{0.64}$ & 47.88 $^{0.54}$ & 54.60 \\
Combine & MCDA \cite{MCD} & 62.91 $^{0.67}$ & 45.77 $^{0.45}$ & 57.57 $^{0.33}$ & 45.88 $^{0.67}$ & 53.03 \\ 
 & \textbf{MUDA} (ours) & {\bf 67.62} $^{0.39}$ & {\bf 52.38} $^{0.20}$ & {\bf 63.41} $^{0.24}$ & {\bf 57.72} $^{0.27}$ & {\bf 60.28} \\
\Xhline{2\arrayrulewidth}
 & DCTN \cite{DCTN} & 62.06 $^{0.60}$ & 48.79 $^{0.52}$ & 58.85 $^{0.55}$ & 48.25 $^{0.32}$ & 54.49  \\ 
Multi-source & M$^3$SDA \cite{M3SDA} & 64.18 $^{0.27}$ & 49.05 $^{0.16}$ & 57.70 $^{0.24}$ & 49.21 $^{0.34}$ & 55.03 \\ 
UDA & MME \cite{MME} & 68.09 $^{0.16}$ & 47.14 $^{0.32}$ & 63.33 $^{0.16}$ & 43.50 $^{0.47}$ & 55.52 \\
 & DAEL \cite{DAEL} & {\bf 69.95} $^{0.52}$ & {\bf 55.13} $^{0.78}$ & {\bf 66.11} $^{0.14}$ & {\bf 55.72} $^{0.79}$ & {\bf 61.73}  \\ 
\Xhline{2\arrayrulewidth}
\end{tabular}
}%
\label{table:mini}
\end{table*}

\medskip
\noindent {\bf Results}
Table \ref{table:msda} and Table \ref{table:mini} show the mean and standard deviation of the classification accuracy after running five random experiments. 
The proposed MUDA achieves 95.74\% (Digits-five) and 60.28\% (miniDomainNet) average accuracy and outperforms the other single-source UDA methods by large margins. 
The tables also compare with recent multi-source methods.
We observe that MUDA presents comparable and often better performance compared to the state-of-the-art multi-source UDA methods. 
Whereas many multi-source methods demand complex training procedures to take advantage of multi-source data, our MUDA can produce competitive results despite the considerably simpler source-combine approach.

\section{Analysis and Discussion}
\label{ssec:discussion}

\begin{figure}[!t]
  \centering
  \begin{minipage}[t]{0.45\linewidth} \centering \footnotesize
  \includegraphics[width=\linewidth]{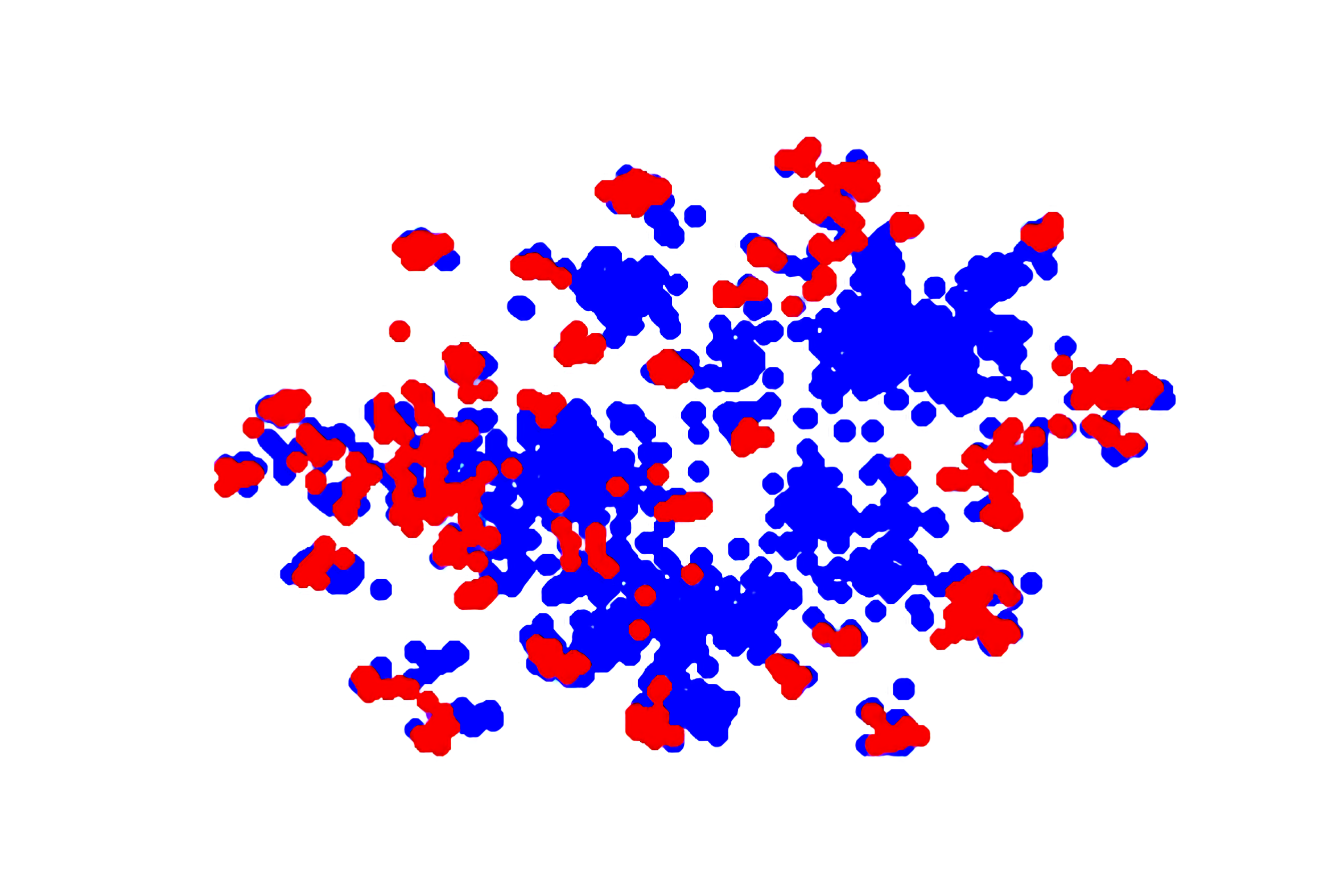}
  (a) Source-only (by domain)
  \end{minipage}
  \begin{minipage}[t]{0.45\linewidth} \centering \footnotesize
  \includegraphics[width=\linewidth]{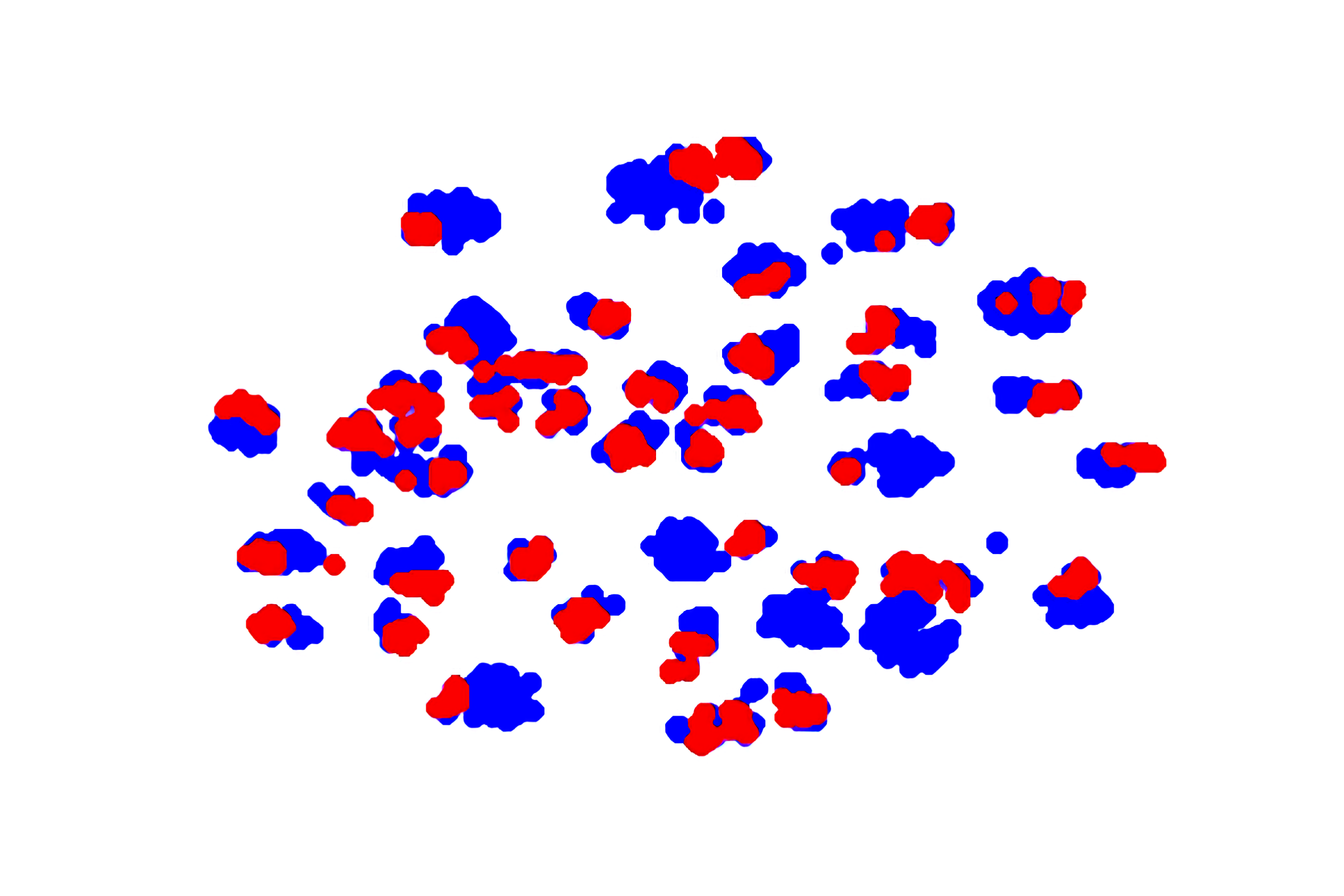}
  (b) MUDA (by domain)
  \end{minipage}
  \medskip
  \begin{minipage}[t]{0.45\linewidth} \centering \footnotesize
  \includegraphics[width=\linewidth]{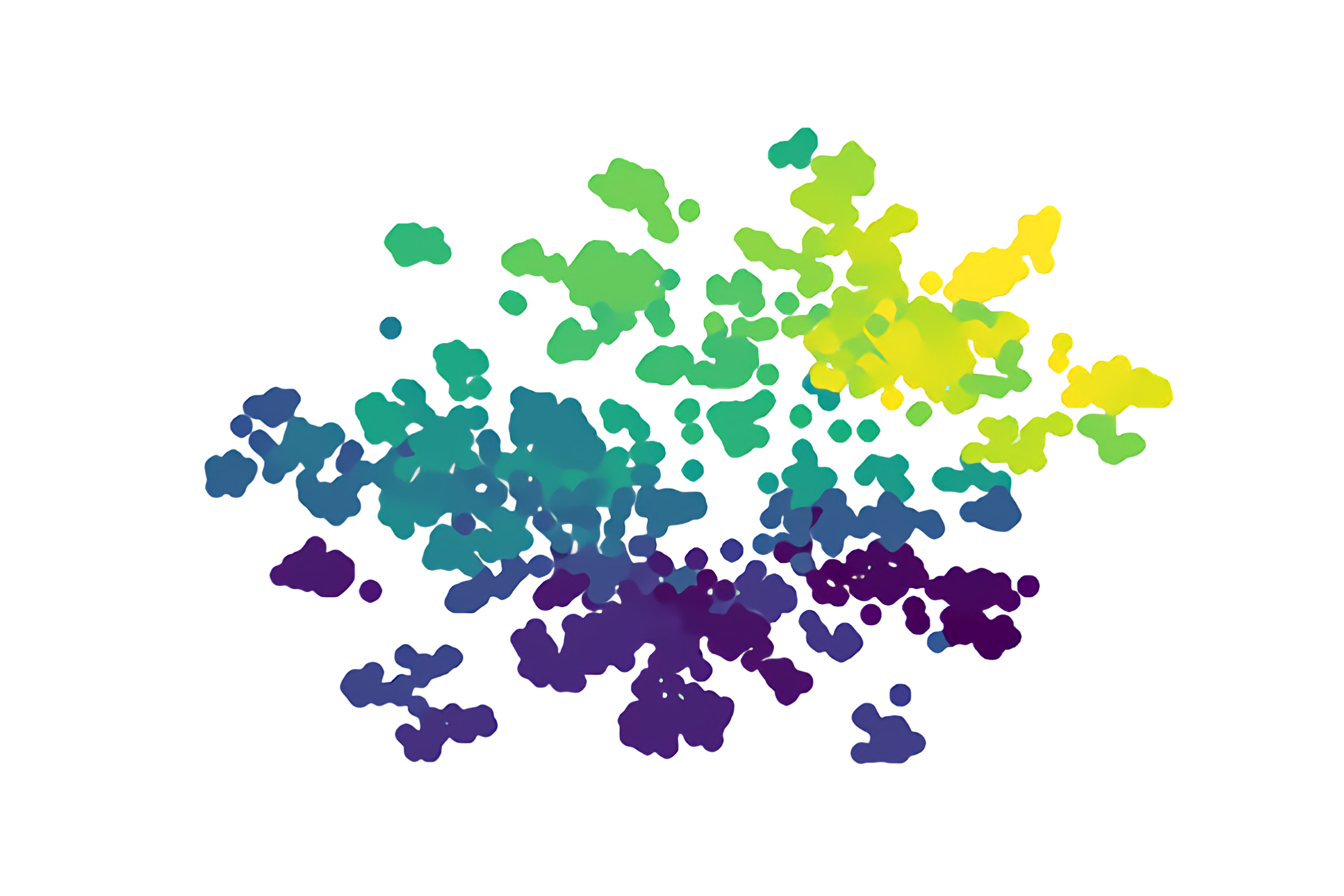}
  (c) Source-only (by class)
  \end{minipage}
  \begin{minipage}[t]{0.45\linewidth} \centering \footnotesize
  \includegraphics[width=\linewidth]{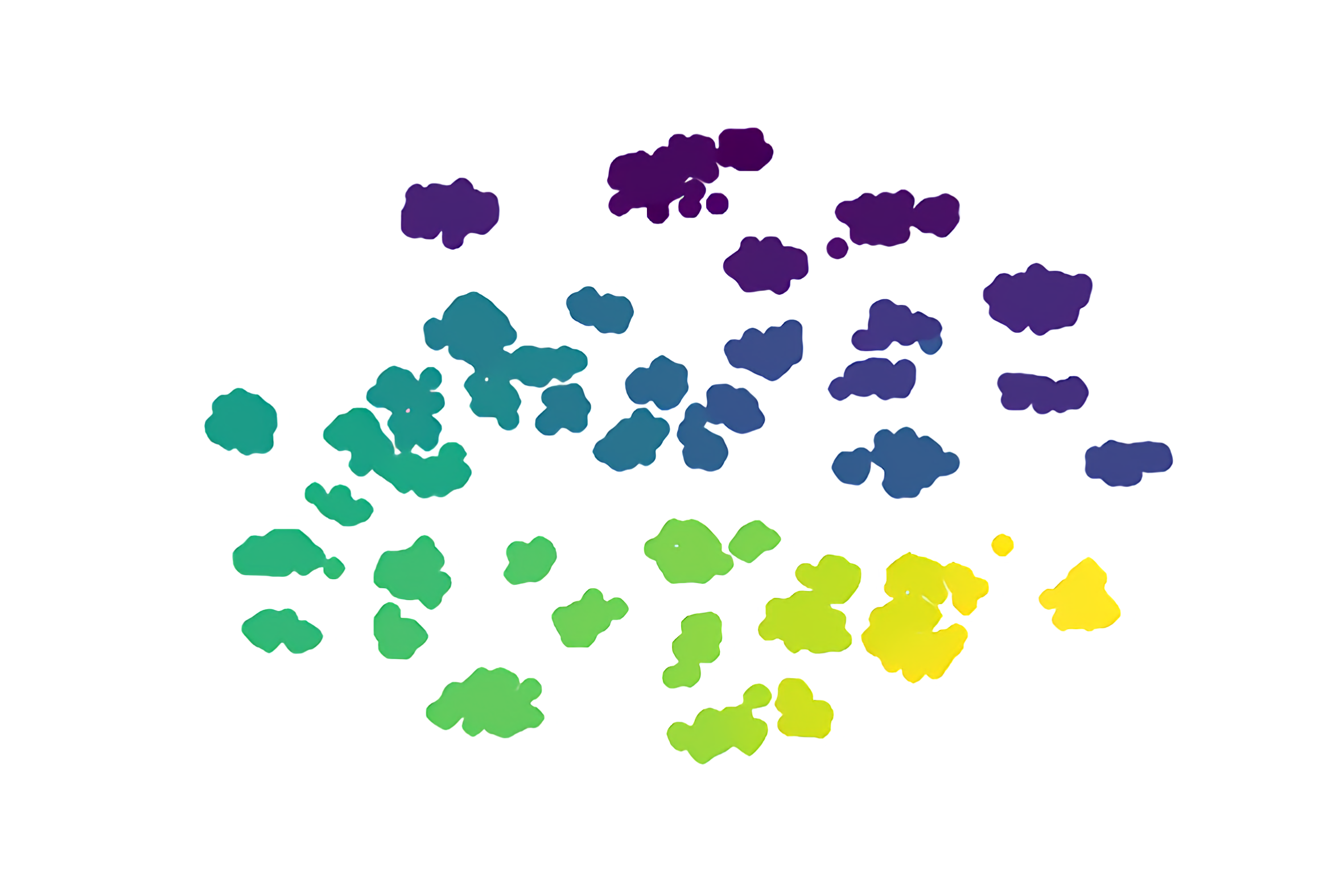}
  (d) MUDA (by class)
  \end{minipage}
\caption{(Best viewed in color) SYNSIG$\rightarrow$GTSRB Analysis. Feature embeddings obtained from the last pooling layer of $F$ are visualized using t-SNE \cite{Maaten08}. Red and blue dots represent the testing samples of the source and the target domain, respectively. We observed that MUDA makes the target samples more discriminative as shown in (b) and (d). 
}
\label{fig:tsne}
\end{figure}

\noindent {\bf Feature Visualization}
In the adaptation from SYNSIG to GTSRB in section \ref{ssec:digit_dataset}, we visualize the learned features using t-SNE \cite{Maaten08}. 
The embeddings in Fig. \ref{fig:tsne} contrast our MUDA with the source-only model. 
Red and blue points in the upper row represent the source and the target data, respectively.
Different colors in the lower row indicate the 43 categories of the target samples. 
The adaptation results in the figure show desirable patterns. 
In Fig. \ref{fig:tsne}b, the feature distributions of both domains align similarly after adaptation. 
This demonstrates that MUDA significantly reduces the divergence between the source and the target domains. 
Furthermore, the adapted features in the same category more tightly group together than those in different categories, as can be seen in Fig. \ref{fig:tsne}d. 
Hence, MUDA makes the target samples become clearer and thus easier to classify. 
This clustering effect is mainly attributed to the model uncertainty minimization approach of MUDA. 
To reduce the predictive variance, MUDA enforces classes so that they contain more consistent features.

\begin{figure}[!t]
\centering
\includegraphics[width=0.7\columnwidth]{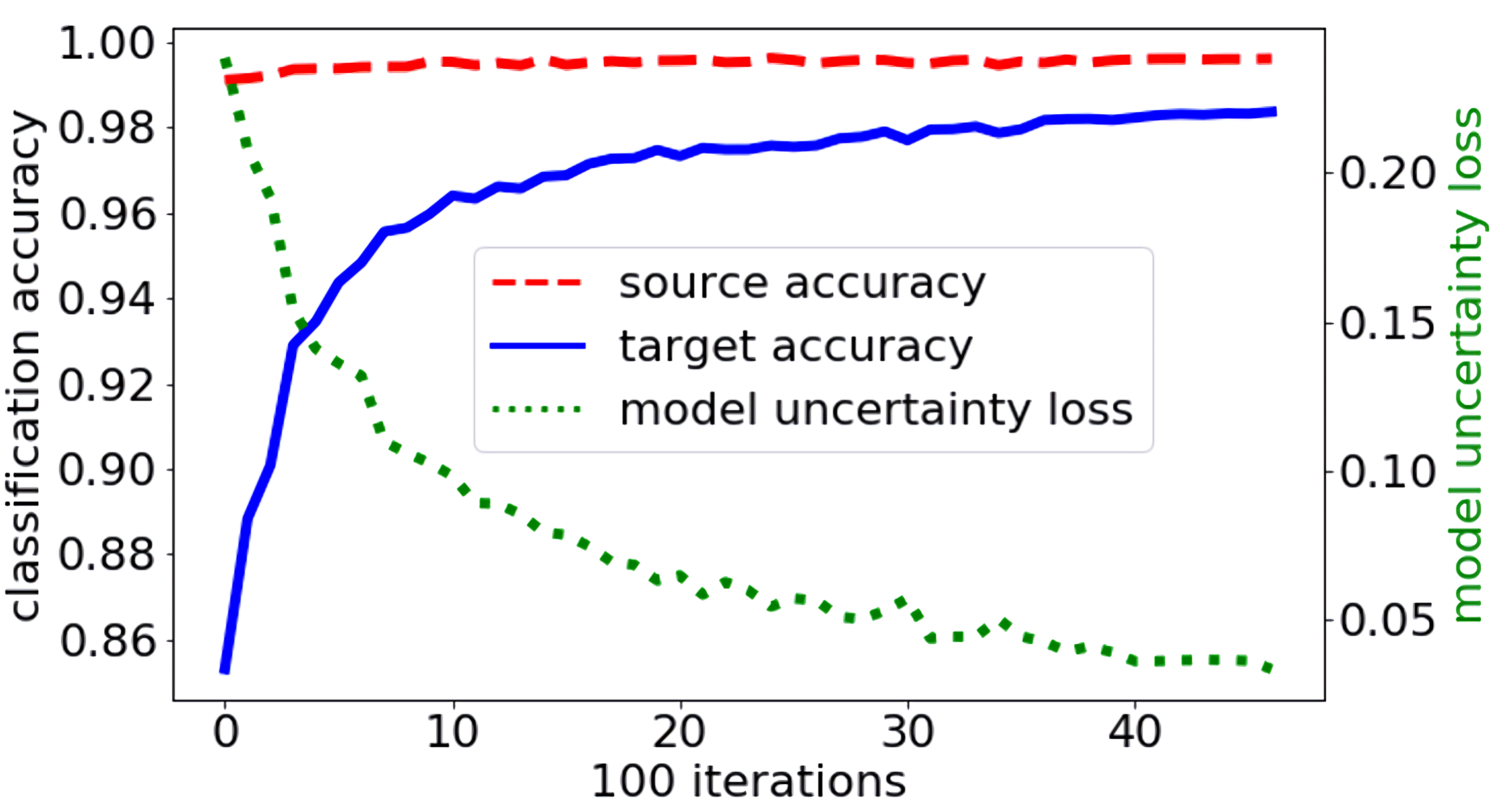}
\caption{(Best viewed in color) MUDA learning curve for SYNSIG to GTSRB. }
\label{fig:convergence}
\end{figure}


\medskip
\noindent {\bf Learning Curve}
For the task SYNSIG $\rightarrow$ GTSRB, we also present the learning curve of MUDA.
Fig. \ref{fig:convergence} depicts how the classification accuracy and the model uncertainly loss change during training. 
As the model uncertainty loss drops rapidly in the early phase and begins to diminish slowly, the target accuracy increases sharply and draws a gentle upward curve. 
In the meantime, the source error is kept small. 
This relationship between the model uncertainty and accuracy confirms that minimizing the model uncertainty on the target samples can improve the domain adaptation outcome.

\begin{figure}[!t]
    \centering
    \includegraphics[width=0.14\linewidth,height=0.14\linewidth]{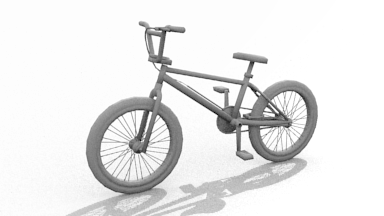}
    \includegraphics[width=0.14\linewidth,height=0.14\linewidth]{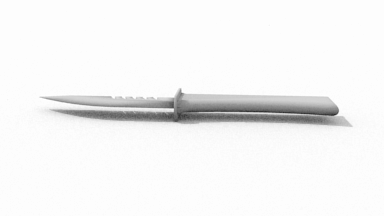}
    \includegraphics[width=0.14\linewidth,height=0.14\linewidth]{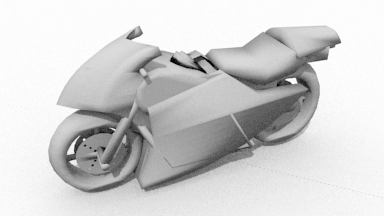}
    \includegraphics[width=0.14\linewidth,height=0.14\linewidth]{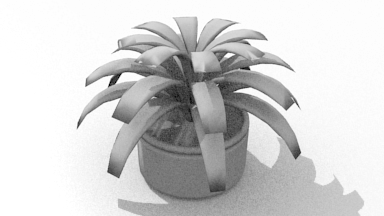}
    \includegraphics[width=0.14\linewidth,height=0.14\linewidth]{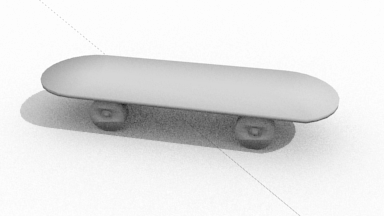}
    \includegraphics[width=0.14\linewidth,height=0.14\linewidth]{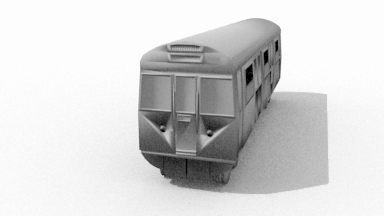}
    \includegraphics[width=0.14\linewidth,height=0.14\linewidth]{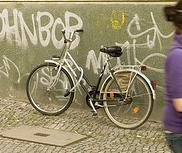}
    \includegraphics[width=0.14\linewidth,height=0.14\linewidth]{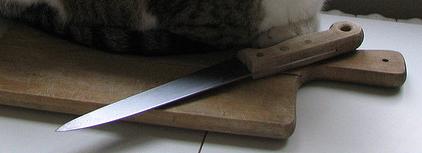}
    \includegraphics[width=0.14\linewidth,height=0.14\linewidth]{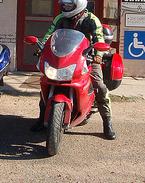}
    \includegraphics[width=0.14\linewidth,height=0.14\linewidth]{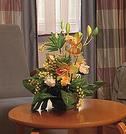}
    \includegraphics[width=0.14\linewidth,height=0.14\linewidth]{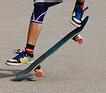}
    \includegraphics[width=0.14\linewidth,height=0.14\linewidth]{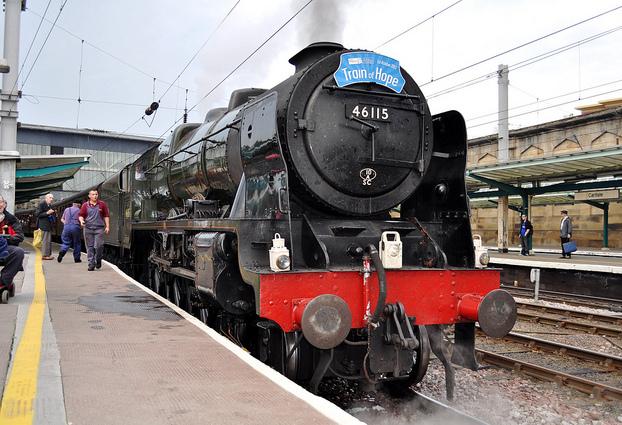}
    \includegraphics[width=0.14\linewidth,height=0.14\linewidth]{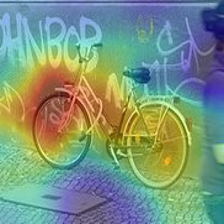}
    \includegraphics[width=0.14\linewidth,height=0.14\linewidth]{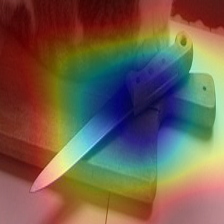}
    \includegraphics[width=0.14\linewidth,height=0.14\linewidth]{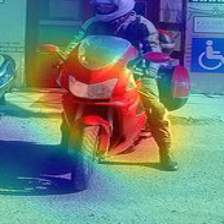}
    \includegraphics[width=0.14\linewidth,height=0.14\linewidth]{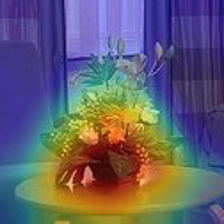}
    \includegraphics[width=0.14\linewidth,height=0.14\linewidth]{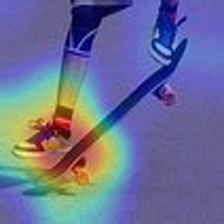}
    \includegraphics[width=0.14\linewidth,height=0.14\linewidth]{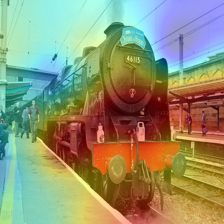}

    \begin{minipage}[t]{0.14\linewidth} \centering 
    \includegraphics[width=\linewidth]{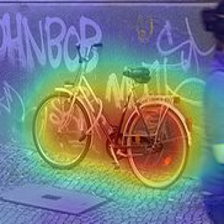}
    (a)
    \end{minipage}
    \begin{minipage}[t]{0.14\linewidth} \centering 
    \includegraphics[width=\linewidth]{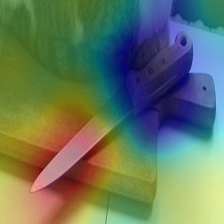}
    (b)
    \end{minipage}
    \begin{minipage}[t]{0.14\linewidth} \centering 
    \includegraphics[width=\linewidth]{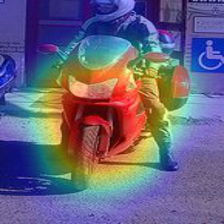}
    (c)
    \end{minipage}
    \begin{minipage}[t]{0.14\linewidth} \centering 
    \includegraphics[width=\linewidth]{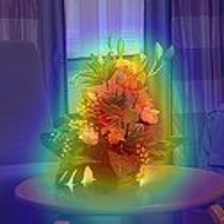}
    (d)
    \end{minipage}
    \begin{minipage}[t]{0.14\linewidth} \centering 
    \includegraphics[width=\linewidth]{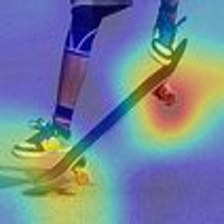}
    (e)
    \end{minipage}
    \begin{minipage}[t]{0.14\linewidth} \centering 
    \includegraphics[width=\linewidth]{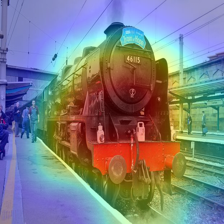}
    (f)
    \end{minipage}

    \caption{(Best viewed in color) Visualization of the class activation map for a prediction in each target image of (a) bicycle, (b) knife, (c) motorcycle, (d) plant, (e) skateboard, and (f) train using Grad-CAM \cite{GradCAM}. The third and the fourth rows are the activation maps before and after MUDA, respectively. The most discriminative regions are emphasized by the deep red color, and the least relevant regions are indicated by the deep blue color.
    }
    \label{fig:gradcam}
\end{figure}

\medskip
\noindent {\bf Class Activation Maps}
To verify that MUDA can effectively capture semantically meaningful features, we examine the class activation maps on several images in the VisDA-17 dataset. 
In Fig. \ref{fig:gradcam}, the activation maps generated by Grad-CAM \cite{GradCAM} highlight the regions that most contribute to the category prediction. 
Red (blue) regions correspond to high (low) scores for class.
The first and the second rows show images of 3D models from the source domain and real images from the target domain, respectively. 
On the target images, the activation maps with or without adaptation are compared.
Grad-CAM overlays from the source-only model are in the third row. 
The adapted results from the proposed MUDA are presented in the last row.

\begin{figure}[!t] 
\centering
  \includegraphics[width=0.8\columnwidth]{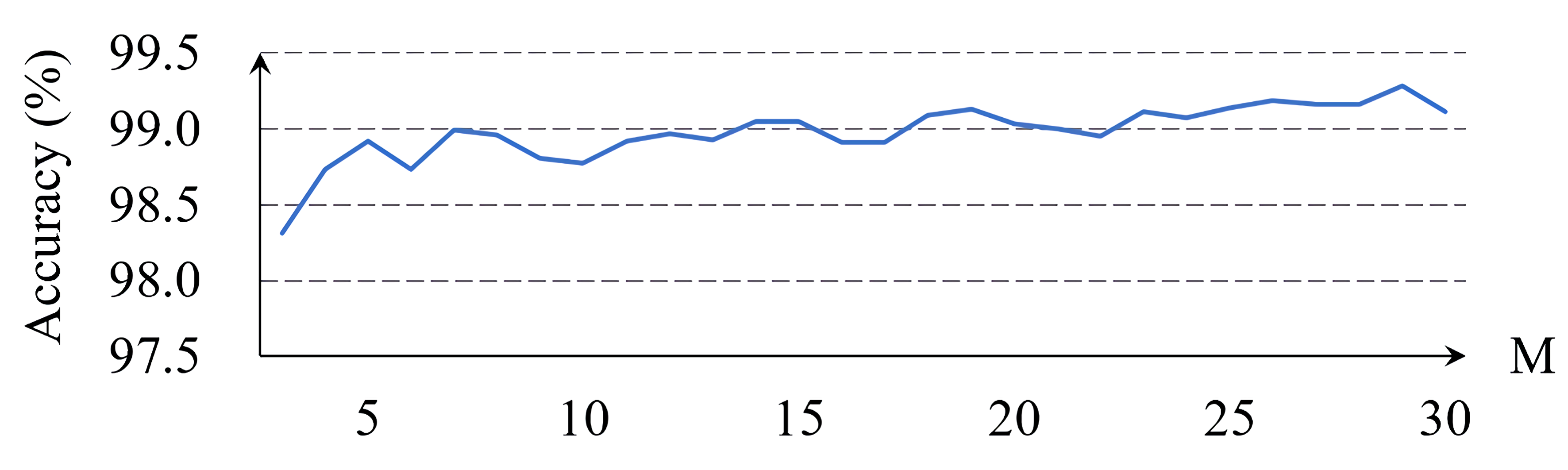}
    \caption{The effect of the number of MC dropout samples on S$\rightarrow$M.}
    \label{fig:analysis_M}
\end{figure}

As shown in the Fig. \ref{fig:gradcam}, the source-only model often fails to locate the relevant regions properly. 
It pays attention to only a portion of the target objects (Fig. \ref{fig:gradcam}a, Fig. \ref{fig:gradcam}e), 
identifies target objects but with less accuracy (Fig. \ref{fig:gradcam}c, Fig. \ref{fig:gradcam}d, Fig. \ref{fig:gradcam}f), 
or even misses the target objects (Fig. \ref{fig:gradcam}b).
On the other hand, the adapted model by MUDA produces more faithful localization maps and focuses on the class-discriminative regions of the target objects. 
These observations indicate that the proposed method MUDA learns domain-invariant features.
Thus, the model can successfully adapt to target samples and generate robust classification results.

\medskip
\noindent {\bf The Choice of $M$}
We investigate how the number of MC dropout samples $M$ affects the performance.
Fig. \ref{fig:analysis_M} shows the change in the classification accuracy on the task SVHN $\rightarrow$ MNIST as $M$ is varied from 3 to 30. 
Though there is some fluctuation, we find that the performance increases slightly as $M$ increases. 
This occurs because the estimate of model uncertainty becomes more accurate with larger values of $M$.
However, we can choose a smaller value for $M$ in practice to reduce the computational burden.

\begin{figure}[!t]
  \centering
  \begin{minipage}[t]{0.48\linewidth} \centering \footnotesize
  \includegraphics[width=\linewidth]{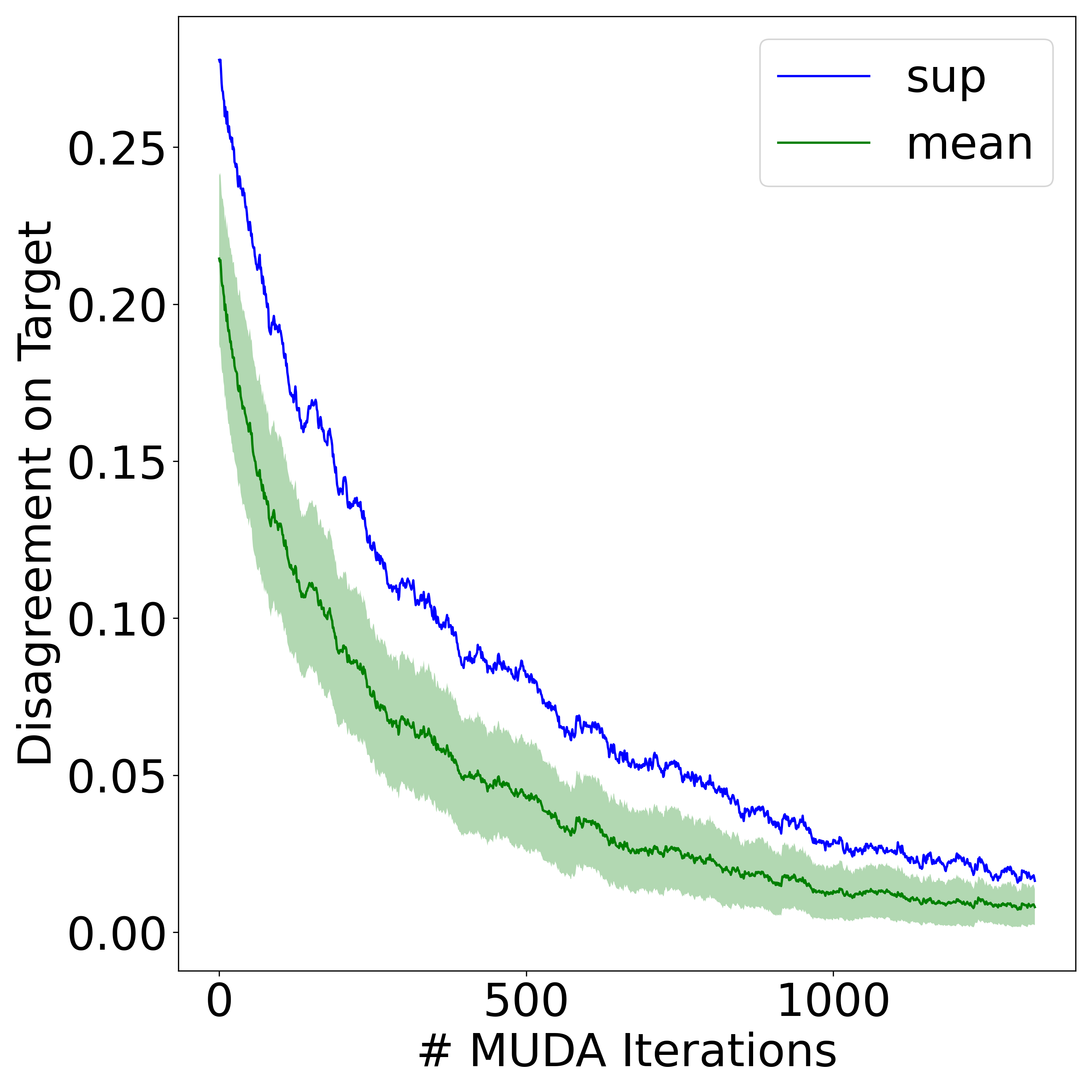}
  (a) Amazon$\rightarrow$DSLR
  \end{minipage}
  \begin{minipage}[t]{0.48\linewidth} \centering \footnotesize
  \includegraphics[width=\linewidth]{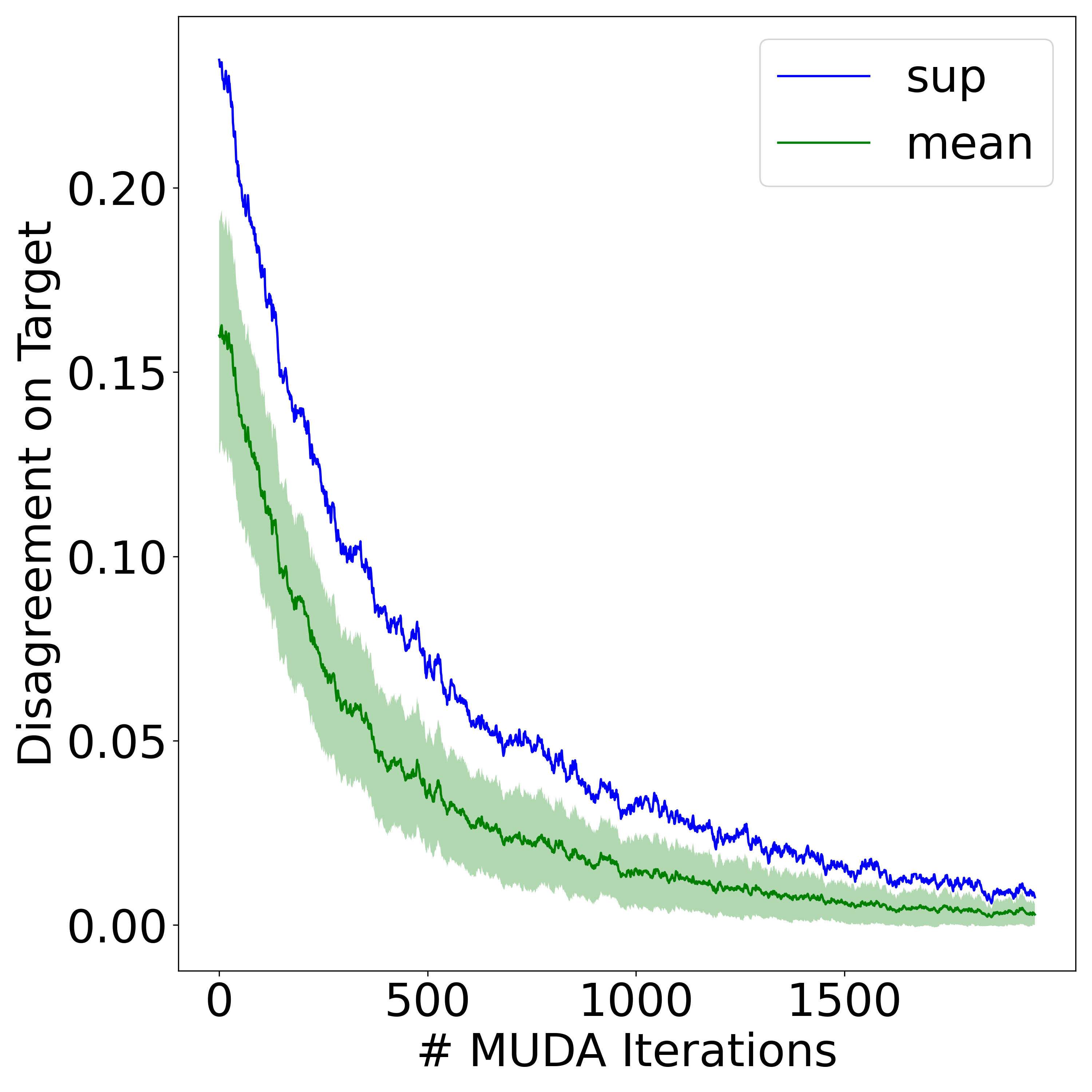}
  (b) Amazon$\rightarrow$Webcam
  \end{minipage}
\caption{
(Best viewed in color) 
We compare the supremum (blue) and the expectation (green) of the target errors of MC dropout samples as MUDA progresses. 
The light green shade represents the standard deviation.
}
\label{fig:risk_analysis}
\end{figure}

\noindent
{\bf Analysis on Supremum and Expectation}
In our formulation of model uncertainty in section \ref{sec:method}, the supremum of the $\H\triangle\H$ divergence is replaced with the expectation. 
We investigate this modification with a numerical analysis.
We compare the supremum of the disagreement rate in equation \ref{eq:app_hsym} with the expected disagreement rate. 
To evaluate these quantities, we compute the disagreement rate $\E_{\D_T}[\ind(h(x)\neq h'(x)]$ between every pair of hypotheses. 
Then, we compare the supremum and the expectation of the target error in this analysis. 
In the computation, MC dropout samples are used. 
Fig. \ref{fig:risk_analysis} illustrates the supremum (blue) and the expectation (green) as MUDA progresses.
Though the supremum is not directly minimized by MUDA, we can observe that the supremum decreases as the expectation decreases. 
The supremum is within the 95\% confidence interval of the expectation all the time, and that their difference gets smaller. 
This empirical analysis supports our intuition and explains why the modified divergence works in practice.

\section{Conclusion}
\label{sec:conc}

In this paper, we presented a novel approach for unsupervised DA. 
Our method generates a domain-adaptive classifier that effectively generalizes to target domain. 
The proposed method uses the Bayesian approach and learns feature representations that reduce the divergence between the source and the target domains by minimizing the model uncertainty.
We demonstrated that our approach outperforms current state-of-the-art methods on challenging image classification benchmarks.

We associated the proposed model uncertainty objective with the classifier-induced domain divergence.
In the derivation, we replaced the supremum in $\H \triangle \H$-divergence with the expectation. 
This reformulation no longer makes the divergence as a valid upper bound of the target error.
However, our intuition and empirical analysis suggest that the target error practically reduces as we minimize this modified divergence. 
We leave a more rigorous theoretical justification as future work.


\appendix
\section{Derivation of Predictive Variance of Hypotheses}
\label{app_a}

Let $\bar{h}(x)=\E_{h \sim \D_\H} [h(x)]$. Then, we can rewrite 
\begin{align}
  & \E_{h,h'\sim \D_\H} [(h(x) - h'(x))^2]  \notag \\
  =~ & \E_{h,h'\sim \D_\H}[(h(x) - \bar{h}(x) + \bar{h}(x) - h'(x))^2] 	\notag \\
	=~ & 2~\E_{h\sim \D_\H}[(h(x) - \bar{h}(x))^2] 
		- 2~\E_{h,h'\sim \D_\H}[(h(x) - \bar{h}(x))(h'(x) - \bar{h}(x))].
\end{align}
Here, the second term vanishes because
\begin{align}
	& \E_{h,h'\sim \D_\H}[(h(x) - \bar{h}(x))(h'(x) - \bar{h}(x))]  \notag \\
	=~ & \E_{h \sim \D_\H}[(h(x) - \bar{h}(x))] 
		~ \E_{h' \sim \D_\H}[(h'(x) - \bar{h}(x))]  \notag \\
	=~ & 0.
\end{align}
Therefore, we have
\begin{align}
  2 ~ \E_{h,h'\sim \D_\H} ~\E_{x \sim \D_T}[(h(x) - h'(x))^2] 
  = 4 ~\E_{x \sim \D_T} ~\E_{h \sim \D_\H} [(h(x) - \bar{h}(x))^2]. 
\end{align}

\bibliography{muda}

\end{document}